\documentclass[journal]{IEEEtran}
\usepackage{amsmath,amsfonts,amssymb,mathrsfs}
\usepackage{algorithm}
\usepackage{algpseudocode} 
\usepackage{array}
\usepackage{textcomp}
\usepackage{stfloats}
\usepackage{url}
\usepackage{verbatim}
\usepackage{graphicx}
\usepackage{makecell}
\usepackage{cite}
\usepackage{multirow}
\usepackage{hyperref}
\usepackage{booktabs}
\usepackage{subcaption}
\usepackage{siunitx}
\usepackage[T1]{fontenc}
\usepackage{textcomp}  
\usepackage[font=small]{caption}
\usepackage[table]{xcolor} 

\hypersetup{
    colorlinks=true,
    linkcolor=blue,
    citecolor=blue,
    urlcolor=blue,
    linkbordercolor={0 0 1}, 
}
\newcommand{\vect}[1]{\boldsymbol{#1}}

\begin{document}
\bstctlcite{BSTcontrol}
\title{Large-Scale Model Enabled Semantic Communication Based on Robust Knowledge Distillation}

\author{Kuiyuan Ding, Caili Guo, ~\IEEEmembership{Senior Member, ~IEEE}, Yang Yang, ~\IEEEmembership{Senior Member, ~IEEE}, Zhongtian Du, 

and Walid Saad, ~\IEEEmembership{Fellow, ~IEEE}

\thanks{K. Ding and Y. Yang are with the Beijing Key Laboratory of Network System Architecture and Convergence, School of Information and Communication Engineering, Beijing University of Posts and Telecommunications, Beijing 100876, China (e-mail: dingkuiyuan@bupt.edu.cn; yangyang01@bupt.edu.cn; wuxiahu@bupt.edu.cn).

C. Guo is with the Beijing Laboratory of Advanced Information Net works, School of Information and Communication Engineering, Beijing University of Posts and Telecommunications, Beijing 100876, China (e-mail: guocaili@bupt.edu.cn).

Z. Du is with the China Telecom Digital Intelligence Technology Company Ltd., Beijing 100035, China (e-mail: duzt@chinatelecom.cn)

W. Saad is with the Bradley Department of Electrical and Computer Engineering, Virginia Tech, Alexandria, VA, 22305 USA (e-mail: walids@vt.edu).

}

\vspace{-1.6em}}


\maketitle


\begin{abstract}
Large-scale models (LSMs) can be an effective framework for semantic representation and understanding, thereby providing a suitable tool for designing semantic communication (SC) systems. However, their direct deployment is often hindered by high computational complexity and resource requirements. In this paper, a novel robust knowledge distillation based semantic communication (RKD-SC) framework is proposed to enable efficient and \textcolor{black}{channel-noise-robust} LSM-powered SC. The framework addresses two key challenges: determining optimal compact model architectures and effectively transferring knowledge while maintaining robustness against channel noise. First, a knowledge distillation-based lightweight differentiable architecture search (KDL-DARTS) algorithm is proposed. This algorithm integrates knowledge distillation loss and a complexity penalty into the neural architecture search process to identify high-performance, lightweight semantic encoder architectures. Second, a novel two-stage robust knowledge distillation (RKD) algorithm is developed to transfer semantic capabilities from an LSM (teacher) to a compact encoder (student) and subsequently enhance system robustness. To further improve resilience to channel impairments, a channel-aware transformer (CAT) block is introduced as the channel codec, trained under diverse channel conditions with variable-length outputs. Extensive simulations on image classification tasks demonstrate that the RKD-SC framework significantly reduces model parameters while preserving a high degree of the teacher model's performance and exhibiting superior robustness compared to existing methods.
\end{abstract}
\begin{IEEEkeywords}
semantic communication, knowledge distillation, neural architecture search, large-scale models.
\end{IEEEkeywords}

\section{Introduction}\label{S1}
Sixth-generation (6G) networks aim to connect trillions of intelligent devices, supporting diverse applications such as augmented reality, medical imaging, and autonomous vehicles\cite{qinzhijinUDEEPSC, 10929033}. However, to achieve this 6G vision, there is a need to address a number of critical challenges, including severe spectrum scarcity and limitations inherent in Shannon’s separate source and channel coding, such as high latency, computational complexity, and suboptimal performance at finite code lengths\cite{sunlunanJSCC, liuchuanhongOFDM}. To overcome these challenges and meet stringent latency and reliability demands, semantic communication (SC) is a promising solution that enables a 6G system to efficiently convey the meaning behind its data rather than transmitting raw data.\cite{10554663}.

SC systems transform raw data into compact semantic representations, which convey the \textit{meaning} of messages\cite{shannon1949mathematical}. Accurate semantic representation is crucial, as it enables SC systems to significantly reduce the amount of transmitted data, save bandwidth resources, and enhance overall communication performance\cite{10554663}, particularly in task-oriented semantic communication (ToSC) scenarios. For improved generalization and robustness of semantic representation, deep learning (DL)-based joint source and channel coding (JSCC) methods have been widely adopted in SC systems\cite{deepJSCC}. Deep neural networks (DNNs) in JSCC are trained via gradient descent to extract semantic information that approximates the minimal sufficient statistics of the raw data, improving robustness against channel-induced interference.

However, existing DL-based JSCC methods often adopt neural networks with limited scale, restricting their semantic representation capabilities. Recent empirical evidence from scaling laws \cite{scalinglaws} indicates that increasing neural network size effectively enhances their capability for semantic representation and understanding thus naturally gives birth to the application of large-scale models in SC systems \cite{qinzhijinUDEEPSC, qin2025lmEmpowerSC, shahid2025largescaleaitelecomcharting}. Recent advancement of artificial intelligence (AI) technologies coupled with significant improvements in computing hardware, particularly graphics processing units (GPUs), large-scale models, represented by large language models (LLMs), emerged as effective learning algorithms that can operate across various general-purpose domains, including natural language understanding, reasoning, and decision-making tasks\cite{zhao2023survey}. LLM frameworks such as DeepSeek-R1 \cite{deepseek-r1}, Grok3 \cite{Grok3}, and chatGPT-o3 \cite{GPT-o3}, can be suitable for designing semantic representation and semantic understanding. 

\subsection{Challenges and Related Works}
\begin{figure*}[t]
    \centering
    \includegraphics[width=.9\linewidth]{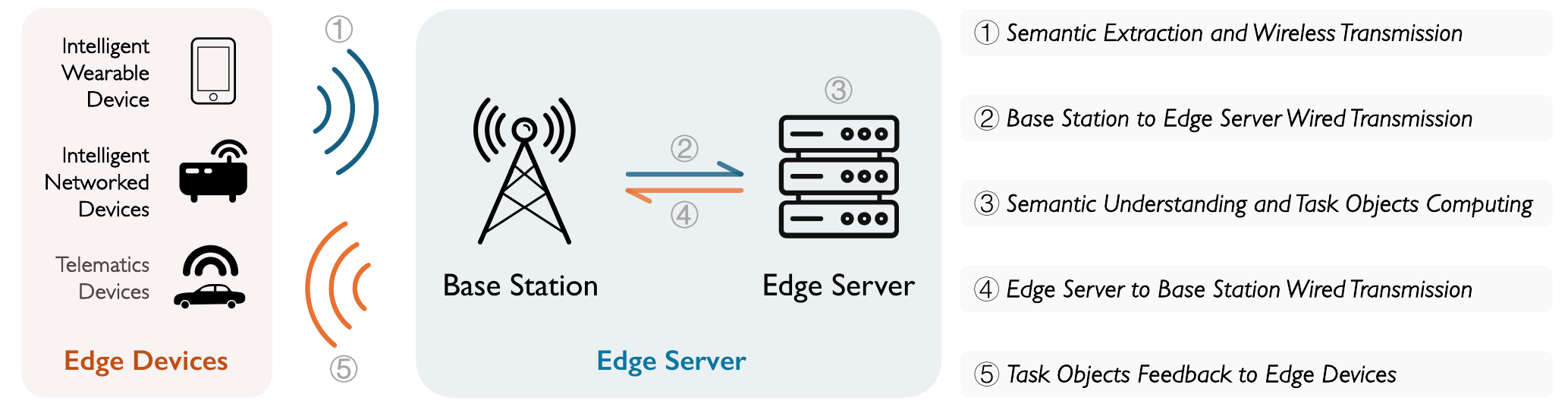}
    \caption{Structure of the considered semantic communication network.}
    \label{fig:structure}
    \vspace{-1em}
\end{figure*}
\subsubsection{Deep JSCC in SC}
A number of recen`t works focused on the application of JSCC in SC systems \cite{farsad2018deeplearningjointsourcechannel, djscc, park2025djscc, xie2021deep, zhenzi2021DeepSC_S, qinzhijin2022multi_user, liuchuanhong2024explainable, fu2024generativeaidriventaskoriented}. These works mainly employ DNNs as the JSCC codec for effective semantic encoding and decoding. 
Specifically, DL-based JSCC was initially introduced in data transmission tasks \cite{djscc, farsad2018deeplearningjointsourcechannel}. In \cite{farsad2018deeplearningjointsourcechannel}, JSCC was applied to sentence embeddings to effectively preserve semantic information. Subsequently, the authors in \cite{djscc} extended the use of JSCC to wireless image transmission, mapping image pixel values to complex-valued channel input symbols, effectively mitigating the ``cliff effec'' inherent in conventional communications. Additionally, an attention-based JSCC framework was proposed in \cite{zhenzi2021DeepSC_S}, utilizing a squeeze-and-excitation network to adapt dynamically to varying channel conditions. Recently, JSCC techniques have been applied to intelligent applications that prioritize task-specific semantic information. For instance, a transformer-based unified transmitter framework for tasks such as image retrieval, machine translation, and visual question answering was proposed in \cite{qinzhijin2022multi_user}. In \cite{liuchuanhong2024explainable}, the authors introduced a triplet-based explainable semantic communication scheme aimed at effectively representing text semantics in text tasks. Furthermore, the work in \cite{fu2024generativeaidriventaskoriented} presented a task-oriented adaptive SC framework employing generative JSCC trained through a generative training algorithm to efficiently transmit task-related semantic features, optimizing bandwidth utilization. Due to limitations in the scale of the DNN approaches in \cite{farsad2018deeplearningjointsourcechannel, djscc, park2025djscc, xie2021deep, zhenzi2021DeepSC_S, qinzhijin2022multi_user, liuchuanhong2024explainable, fu2024generativeaidriventaskoriented}, these existing solutions exhibit constrained semantic representation capabilities, resulting in JSCC codecs typically tailored only to specific datasets. This limitation conflicts with the generalization performance in multiple data scenarios required by practical communication systems. \textcolor{black}{Large-scale models, with their vast parameter counts and exposure to diverse training data, inherently possess the powerful and generalizable ability of semantic representation needed to overcome this challenge.} Therefore, exploring how to effectively integrate large-scale models within semantic communication systems remains an important open research issue.

\subsubsection{Large-Scale AI Models for SC}
A number of recent works studied the use of large-scale models, particularly LLMs, to enhance semantic representation and semantic understanding. These studies primarily focus on semantic encoding and decoding within SC systems \cite{wang2024largelanguagemodelenabled, ribouh2025largelanguagemodelbasedsemantic, guoshuaishuai2025, yang2025rethinkinggenerativesemanticcommunication, xiang2025sceneunderstandingenabledsemantic}. Specifically, in \cite{wang2024largelanguagemodelenabled}, the authors proposed an LLM-enabled semantic communication framework, applying LLMs directly to physical layer coding and decoding for text transmission. The authors in \cite{ribouh2025largelanguagemodelbasedsemantic} introduced an orthogonal frequency-division multiplexing (OFDM)-based semantic communication framework for image transmission, exploiting the cross-modal understanding capabilities of LLMs for efficient encoding and decoding. In \cite{guoshuaishuai2025}, the authors proposed the use of LLMs to quantify semantic importance and perform error correction in semantic representations of raw visual data. 
While these studies primarily employed LLMs to enhance data transmission processes, several other investigations have explored the use of LLMs in performing intelligent tasks. In \cite{yang2025rethinkinggenerativesemanticcommunication}, the authors presented a novel generative semantic communication framework for 6G multi-user systems based on multi-modal large language models (MLLMs), which serve as a shared knowledge base facilitating standardized semantic encoding and personalized decoding. In \cite{xiang2025sceneunderstandingenabledsemantic}, the authors developed an innovative OpenSC system for 6G semantic communications by integrating scene understanding, LLMs, and open channel coding techniques. This approach enables adaptive and generalizable semantic encoding, significantly enhancing transmission efficiency and overcoming the limitations posed by static coding and task-specific knowledge bases in traditional SC systems. 

While these prior works \cite{wang2024largelanguagemodelenabled, ribouh2025largelanguagemodelbasedsemantic, guoshuaishuai2025, yang2025rethinkinggenerativesemanticcommunication, xiang2025sceneunderstandingenabledsemantic} have demonstrated the advantages of employing large-scale models in SC systems, they rarely address the potential drawbacks such as high codec delays and substantial computing resource demands. These factors are critical since they determine the practicality and feasibility of deploying such methods in real-world communication systems. Hence, compressing large-scale models to satisfy delay and energy consumption constraints becomes essential. Knowledge distillation (KD) \cite{hinton2015distillingknowledgeneuralnetwork} is an effective method to solve this problem by compressing the large-scale model (teacher model) into a smaller one (student model) through knowledge transfer. Consequently, KD can be used to generate a compact, yet highly performant semantic encoder for affordable large-scale model enabled SC. However, there are two primary challenges that must be addressed:

\begin{figure*}[t]
    \centering
    \includegraphics[width=0.9\linewidth]{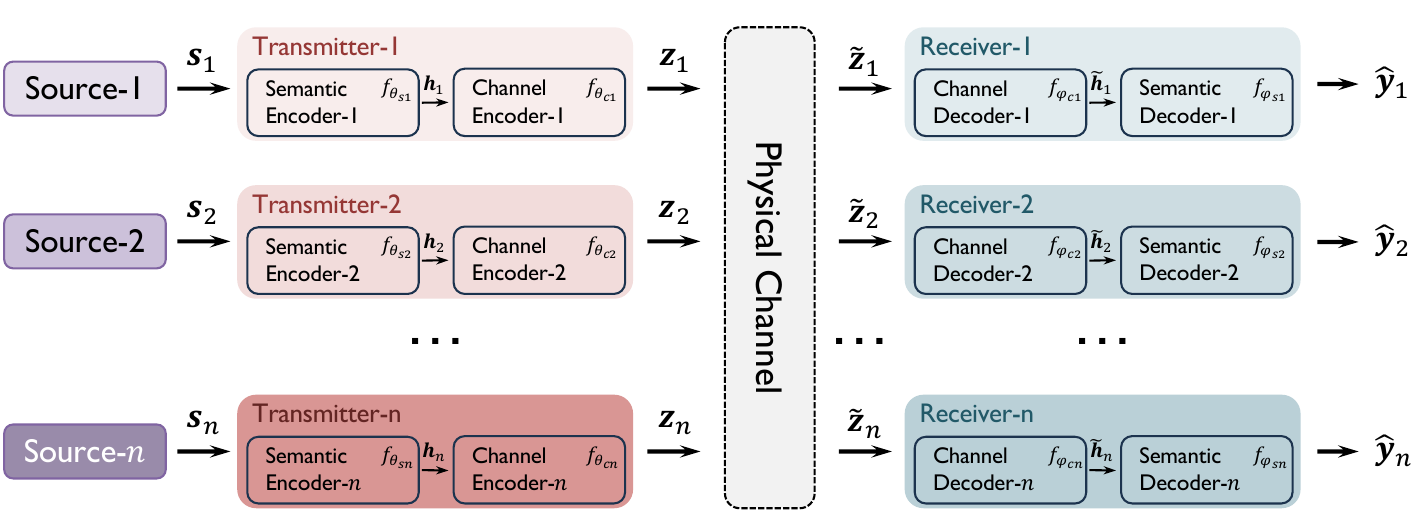}
    \caption{Illustration of the considered system model.}
    \label{fig:system-model}
\end{figure*}

\begin{itemize}
\item \textit{Challenge 1: How can we determine optimal small-scale model architectures that effectively balance the learning capacity from the teacher model and computational complexity across diverse application scenarios?}
\item \textit{Challenge 2: How can KD compress large-scale models while preserving semantic representation-understanding capabilities and robustness against channel noise?}
\end{itemize}

\subsection{Contributions}
The main contribution of this paper is the development of a robust knowledge distillation-based semantic communication (RKD-SC) framework\footnote{An earlier version of this work, presenting preliminary results on the RKD-SC framework, was published in the IEEE Global Communications Conference (GLOBECOM) 2024 \cite{RKDSC}.}, which integrates a knowledge distillation-based lightweight differentiable architecture search algorithm (KDL-DARTS) with a novel two-stage robust knowledge distillation algorithm (RKD). To the best of our knowledge, this is the first work combining neural architecture search (NAS) and knowledge distillation (KD) to distill large-scale model intelligence into a compact semantic encoder optimized for robust semantic feature transmission. Specifically, our contributions include:

\begin{itemize}
    \item We propose the RKD-SC framework, which first employs KDL-DARTS to address Challenge 1. As an extension of the differentiable architecture search (DARTS) framework \cite{liu2018darts}, KDL-DARTS learns a set of continuous variables to weight the outputs of candidate operations guided by the knowledge distillation loss. Additionally, it introduces a penalty factor to encourage the selection of operations with fewer parameters. The proposed algorithm effectively searches for optimal compact architectures that achieve an optimal trade-off between task performance and model complexity. 
    
    \item Subsequently, within the RKD-SC framework, we utilize RKD to tackle Challenge 2 which involves at transferring knowledge from a large-scale model to a compact semantic encoder. Specifically, the first stage of RKD emphasizes enhancing semantic representational capabilities, whereas the second stage specifically focuses on robustness enhancement.
    
    \item Considering the degradation in robustness typically associated with lightweight models, we introduce the channel-aware transformer (CAT) to improve the ability of the RKD-SC system against channel noise. The proposed CAT is trained under diverse channel conditions and employs variable-length output dimensions to effectively balance data throughput and robustness.
\end{itemize}

Simulation results show that the proposed RKD-SC framework preserves 95.86\% of the performance of the teacher model while reducing the number of parameters by approximately 94.06\% at an SNR of 25 dB and achieves performance gains exceeding 83.12\% compared to the teacher model at an SNR of $-$10 dB on CIFAR10 dataset. Our results also demonstrate that the RKD-SC framework effectively transfers capabilities from large-scale to small-scale models while simultaneously maintaining robustness.

The rest of the this paper is organized as follows. Section \ref{S2} outlines the system model. Section \ref{S3} details the proposed RKD-SC framework including KDL-DARTS algorithm and RKD algorithm. Section \ref{S4} presents comprehensive experimental evaluations to validate the effectiveness of our proposed framework. Finally, Section \ref{S5} concludes the paper.

\section{System Model and Problem Formulation}\label{S2}

We consider a distributed SC system, as illustrated in Fig. \ref{fig:structure}. Initially, edge devices such as smart wearables, intelligent networked devices and telematics units extract semantic features and transmit them wirelessly to the base station. Subsequently, the base station forwards the received noisy semantic features to the edge server over a wired link. Upon reception, the edge server interprets these semantic features and decodes them into task-specific objects. After computation, the processed task-specific objects are transmitted back from the edge server to the base station. Finally, the base station provides feedback regarding these task-specific objects to the edge devices, enabling real-time adaptation and response. Next, we first introduce the SC system model illustrated in Fig. \ref{fig:system-model}. Then, we formally present the optimization problem formulated to address Challenge 1 and Challenge 2.


\subsection{System Model}
As shown in Fig. \ref{fig:system-model}, the considered system consists of multiple transmitters such as edge devices or sensors, physical wireless channels, and multiple signal receivers hosted on an edge server. 
For the $i$-th transmitter, a semantic encoder, denoted as $f_{\boldsymbol{\theta}_{si}}$ parameterized by ${\boldsymbol{\theta}_{si}}$, encodes the source message $\boldsymbol{s}_i$ into a compact semantic feature:
\begin{equation}
\label{eq1-1}
    \boldsymbol{h}_i = f_{\boldsymbol{\theta}_{si}}\left(\boldsymbol{s}_i\right).
\end{equation}

This semantic feature $\boldsymbol{h}_i$ is subsequently encoded by a channel encoder, $f_{\boldsymbol{\theta}_{ci}}$, parameterized by ${\boldsymbol{\theta}_{ci}}$, to yield the transmitted symbol $\boldsymbol{z}_i$, which can be expressed as:
\begin{equation}
\label{eq1}
    \boldsymbol{z}_i =  f_{\boldsymbol{\theta}_{ci}}\left(\boldsymbol{h}_i\right).
\end{equation}

The transmitted symbol $\boldsymbol{z}_i$ is sent through the physical channel, and the symbol received at the receiver is given by:
\begin{equation}
    \label{eq2}
    \tilde{\boldsymbol{z}}_i = \boldsymbol{H}\boldsymbol{z}_i + \boldsymbol{n},
\end{equation}
where $\boldsymbol{H} \in \mathbb{C}$ is the channel gain coefficient, and $\boldsymbol{n} \sim \mathcal{CN}\left( \mathbf{0}, \sigma^2 \right)$ represents the additive white Gaussian noise (AWGN).

At the receiver, a single channel decoder and semantic decoder process the received symbols from all transmitters. Specifically, the $i$-th channel decoder, denoted by the $f_{\boldsymbol{\varphi}_{ci}}$ and parameterized by ${\boldsymbol{\varphi}_{ci}}$, decodes the received symbol $\tilde{\boldsymbol{z}}_i$ into the following semantic feature:

\begin{equation}
\label{eq3-1}
\tilde{\boldsymbol{h}}_i = f_{\boldsymbol{\varphi}_{ci}}\left( \tilde{\boldsymbol{z}}_i \right).
\end{equation}

The decoded semantic feature $\tilde{\boldsymbol{h}}_i$ is then processed by the $i$-th semantic decoder $f_{\boldsymbol{\varphi}_{si}}$, parameterized by $\boldsymbol{\varphi}_{si}$, to obtain the following task target:

\begin{equation}
\label{eq3}
    \hat{\boldsymbol{y}}_i = f_{\boldsymbol{\varphi}_{si}}\left( \tilde{\boldsymbol{h}}_i \right).
\end{equation}

In this SC system, the key goal is to identify optimal architectures for semantic encoders that effectively balance performance and computational complexity while maintaining semantic representation-understanding capabilities and robustness of the whole system.


\subsection{Problem Formulation}
While the use of large-scale models can be effective for semantic representation, their direct applicability in real-world devices can be impractical due to the associated computational overhead. To reduce computational overhead and latency while maintaining the capabilities of the large-scale model, we seek to transfer knowledge from a large-scale model to a smaller-scale one serving as a semantic encoder by KD. 
To achieve this goal, two primary challenges must be addressed as discussed in Section \ref{S1}, which can be captured by the following optimization problem:
\begin{equation}
\label{total_op}
    \left\{a_i^*,\boldsymbol{\theta_{si}}^*\right\} = \arg\max_{a_i,\boldsymbol{\theta_{si}}} \Bigl[\mathbb{E} \bigl(\mathcal{R}(a_i,\boldsymbol{\theta_{si}})\bigr)\Bigr],
\end{equation}
where $a_i^*$ represents the optimal architecture corresponding to Challenge 1 and $\boldsymbol{\theta_{si}}^*$ is the vector of optimal network parameters corresponding to Challenge 2. $\mathcal{R}$ represents a performance metric employed to evaluate the system’s effectiveness, which will be explicitly defined below. The notation $\mathbb{E}(\cdot)$ is the expectation operation.

There are two optimization objectives embedded within problem (\ref{total_op}): determining the optimal neural architecture and finding the optimal network parameters. Consequently, the original optimization problem (\ref{total_op}) can be further decomposed into two distinct sub-problems: (a) a neural architecture search (NAS) problem aimed at identifying the optimal architecture and (b) a knowledge distillation (KD) problem aimed at determining optimal network parameters guided by a large-scale model. In the following, we first formally describe the objective of the NAS sub-problem, followed by the objective formulation of the KD sub-problem.

\subsubsection{NAS Objective}
Challenge 1 can be addressed by formulating it as a neural architecture search problem. Considering the semantic encoder $f_{\boldsymbol{\theta_{si}}}$ at transmitter $i$, we define $\boldsymbol{\mathcal{A}}_i$ as the search space containing all candidate student model architectures of $f_{\boldsymbol{\theta_{si}}}$, where each architecture $a_i^{(k)} \in \boldsymbol{\mathcal{A}}_i$ corresponds to a particular model configuration (e.g., the number of layers and channels). Let $\mathcal{S}_{a_i^{(k)},\boldsymbol{\theta_{si}}}$ denote the student model (i.e., semantic encoder $f_{\boldsymbol{\theta_{si}}}$) with architecture $a_i^{(k)}$ and trainable parameters $\boldsymbol{\theta_{si}}$. Our goal is to identify optimal compact model architectures that effectively balance the learning capability derived from the teacher model and computational complexity across diverse application scenarios. Specifically, we define the following components:

\begin{itemize}
    \item \emph{Performance measure}: For a given architecture $a$ and parameters $\boldsymbol{\theta_{si}}$, $\mathcal{P}(a_i^{(k)},\boldsymbol{\theta_{si}})$ is the performance metric (e.g., accuracy or KD loss) of the student model evaluated on a standard validation or test set without noise interference.
    \item \emph{Model complexity}: $\Omega(a_i^{(k)})$ is the complexity of architecture $a$, represented by a function $\Omega(\cdot)$ of the number of normalized parameters.
\end{itemize}

To jointly consider performance and computational cost, we define a comprehensive optimization objective as follows:
\begin{equation}
\label{eq4}
    \mathcal{R}(a_i^{(k)},\boldsymbol{\theta_{si}}) = \eta \,\mathcal{P}(a_i^{(k)},\boldsymbol{\theta_{si}}) - \zeta\,\Omega(a_i^{(k)}),
\end{equation}
where $\eta$ and $\zeta$ are positive hyperparameters that determine the relative importance of each term.=

By addressing the Challenge 1—searching for a suitable yet compact model architecture—within the NAS framework, our objective becomes finding an optimal architecture $a_i^*$ that maximizes the expected optimization objective $\mathcal{R}(a_i^{(k)},\boldsymbol{\theta_{si}})$. Formally, the NAS optimization involves evaluating a substantial number of candidate architectures to identify $a^*$.

Upon convergence of the search process, the architecture yielding the highest expected $\mathcal{R}$ is selected:
\begin{equation}
\label{eq5}
    a_i^* = \arg\max_{a_i^{(k)} \in \mathcal{A}_i} \Bigl[\mathbb{E}_{\boldsymbol{\theta_{si}} \sim \text{Train}(a_i^{(k)}\mid \mathcal{D}_i)} \bigl(\mathcal{R}(a_i^{(k)},\boldsymbol{\theta_{si}})\bigr)\Bigr],
\end{equation}
where $\boldsymbol{\theta_{si}} \sim \text{Train}(a_i^{(k)}\mid \mathcal{D}_i)$ denotes the model parameters $\boldsymbol{\theta_{si}}$ obtained after training architecture $a$ on the dataset $\mathcal{D}_i$.

\subsubsection{KD Objective}
After determining the optimal architecture $a_i^*$, we address Challenge 2 by refining the parameters $\boldsymbol{\theta_{si}}$ of the selected architecture using KD.

Specifically, the student model $S_{a_i^*,\boldsymbol{\theta_{si}}}$ is distilled from a teacher model $f_{\boldsymbol{\theta_{t}}}$, parameterized by $\boldsymbol{\theta_{t}}$. The goal is for the student model to acquire semantic representation capabilities comparable to the teacher while maintaining robustness against channel noise. Given a dataset $\mathcal{D}_i=\left\{(\boldsymbol{x},\boldsymbol{y})\right\}$ consisting of samples $\boldsymbol{x}$ and corresponding labels $\boldsymbol{y}$, we denote the semantic features extracted by the teacher and student models as $\boldsymbol{h}_i^\text{Tea}$ and $\boldsymbol{h}_i$, respectively. These can be formulated as:

\begin{equation}
\label{eq7}
    \boldsymbol{h}_i^\text{Tea}= f_{\boldsymbol{\theta_{t}}}(\boldsymbol{x}) \quad \text{and} \quad \boldsymbol{h}_{i} = S_{a_i^*,\boldsymbol{\theta_{si}}}(\boldsymbol{x}).
\end{equation}

To preserve the semantic representation capabilities of the teacher, the student model is trained to minimize the discrepancy between $\boldsymbol{h}_i$ and $\boldsymbol{h}_i^\text{Tea}$, formulated as:

\begin{equation}
    \label{eq8}
    \boldsymbol{\theta_{si}^*} = \arg\min_{\boldsymbol{\theta_{si}}} \mathcal{L}_{\text{KD}}\left( \boldsymbol{h}_i^\text{Tea}, \boldsymbol{h}_i \right),
\end{equation}
where $\mathcal{L}_{\text{KD}}$ measures the semantic feature difference between the large-scale (teacher) model and the small-scale (student) model. Here, $\boldsymbol{\theta_{si}^*}$ represents the optimal student parameters minimizing $\mathcal{L}_{\text{KD}}$.

Considering the decoding phase, we aim for the semantic representation $\boldsymbol{h}_i^\text{Tea}$ to be accurately decoded by the semantic decoder $f_{\boldsymbol{\varphi_{si}}}$ with channel noise $\boldsymbol{n}$. Thus, an additional loss term that compares student model outputs to ground truth labels should be incorporated into (\ref{eq8}), leading to:
\begin{equation}
    \label{eq9}
    \boldsymbol{\theta_{si}^*}, \boldsymbol{\varphi_{si}^*} = \arg\min_{\boldsymbol{\theta_{si}},\boldsymbol{\varphi_{si}}} \lambda\mathcal{L}_{\text{KD}}\left( \boldsymbol{h}_i^\text{Tea}, \boldsymbol{h}_i \right) 
    + (1-\lambda)\mathcal{L}_{\text{task}}\left(\boldsymbol{\hat{y}}_i, \boldsymbol{y}_i \mid \boldsymbol{n}\right),
\end{equation}
where $\boldsymbol{\hat{y}}_i$ is the receiver output when the transmitter sends the semantic representation $\boldsymbol{h}_i$ through the physical channel. $\mathcal{L}_\text{task}$ is the task loss function, which measures the discrepancy between the system output $\boldsymbol{\hat{y}}_i$ and the task ground truth $\boldsymbol{y}_i$, $\lambda$ is a positive hyperparameter that determine the relative importance of each term.
 
This two-stage process (NAS for architecture selection, followed by KD on the chosen architecture) yields a high-performing, compact, and more robust student model $\mathcal{S}_{a_i^*,\boldsymbol{\theta_{si}^*}}$, thereby addressing both challenges highlighted in Section~\ref{S1}. 

As discussed in the subsequent sections, to optimize these two objective, we propose a RKD-SC framework which utilizes the KDL-DARTS algorithm to obtain $a_i^*$ and utilizes KD to compress the large-scale model into the smaller-scale model architected by $a_i^*$ for semantic encoding, and enhances the robustness of the system against channel noise with CAT. 

\section{Proposed RKD-SC Framework}\label{S3}

In this section, we present the proposed RKD-SC framework, where a high-performance and compact semantic encoder architecture is initially identified using the KDL-DARTS algorithm. Subsequently, the overall system robustness is enhanced through the proposed RKD algorithm, and the CAT module functions as the channel codec. In what follows, we first introduce the KDL-DARTS algorithm, then describe the architecture of the proposed CAT module, and finally present the RKD algorithm in detail.

\subsection{KD-based Lightweight DARTS}

\begin{figure*}[t!]
    \centering
    \includegraphics[width=.75\linewidth]{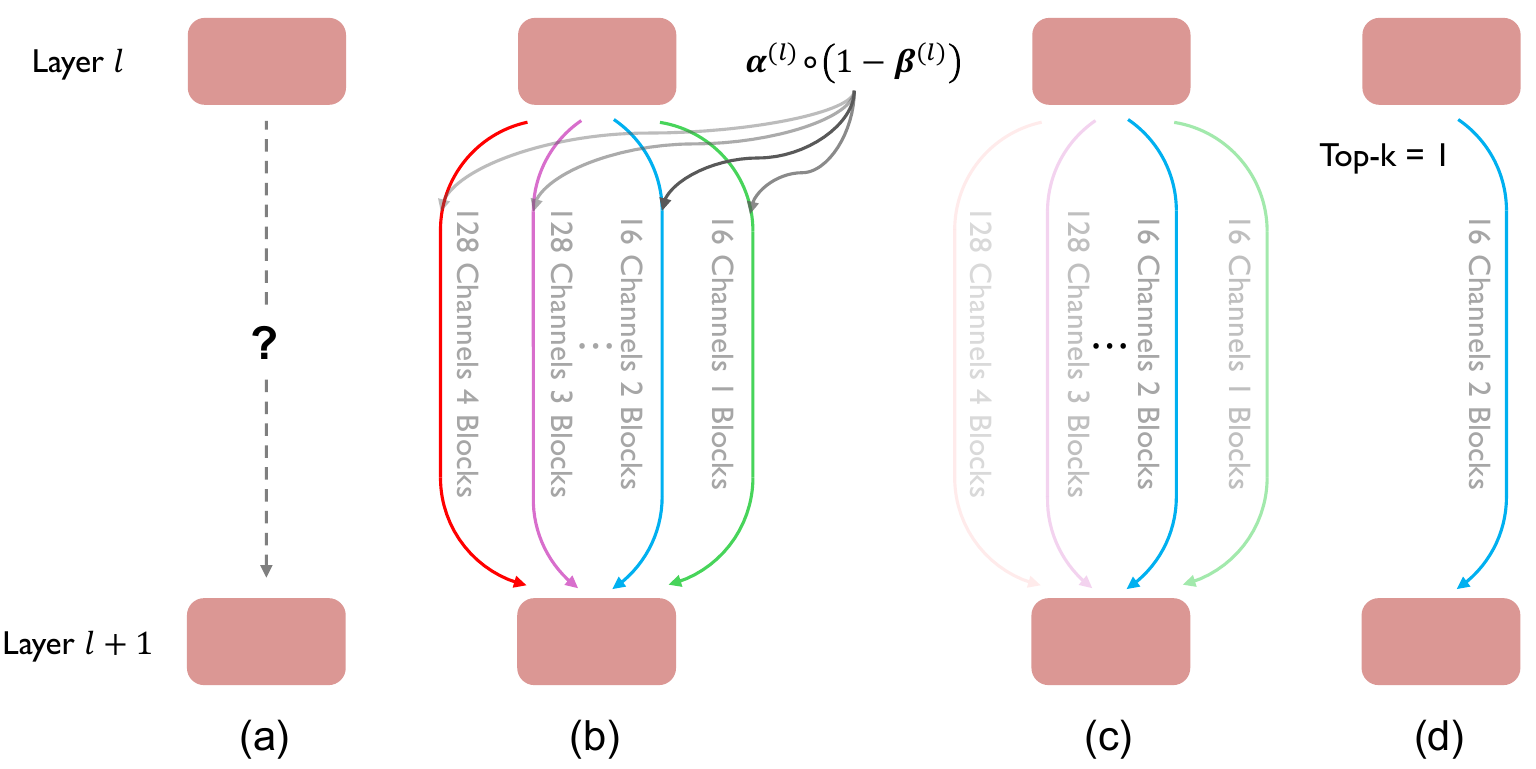}
    \caption{An overview of the proposed KDL-DARTS method: (a) The operations between layer $l$ and layer $l+1$ are initially undetermined. (b) Candidate operations are introduced, and their optimal continuous operation weights, denoted as $\boldsymbol{\alpha}^{(l)}$, are derived by solving a bilevel optimization problem. (c) A combined metric is computed using the Hadamard product: $\boldsymbol{\alpha}^{(l)} \circ (1 - \boldsymbol{\beta}^{(l)})$. (d) The top-$k$ optimal operations are selected based on the values obtained from this combined metric.}
    \label{fig:KDL-DARTS}
\end{figure*}

\textcolor{black}{To solve the NAS sub-problem in Eq.~\eqref{eq5}, we first employ our proposed KDL-DARTS algorithm. Its goal is to identify a computationally efficient architecture for the semantic encoder ($f_{\boldsymbol{\theta_{si}}}$) that excels at learning distilled semantic representations. Following DARTS~\cite{liu2018darts}, KDL-DARTS searches over a predefined set of candidate operations (e.g., convolutional or attention blocks) to construct the optimal encoder architecture.}
Specifically, each operation $o^{(l,j)}\in \mathcal{O}^{(l)}$ is applied to the input $x^{(l)}$ of the $l$-th layer, and is defined as:

\begin{equation}
    y^{(l,j)} = o^{(l,j)}( x^{(l)} ),
\end{equation}
where $y^{(l,j)}$ is the output generated by the operation $o^{(l,j)}$.

Consistent with the DARTS framework, we introduce a vector of continuous candidate operation weights $\boldsymbol{\alpha}^{(l)}=\{\alpha^{(l,j)}\}\subsetneq \boldsymbol{\alpha} $ to combine outputs from all candidate operations between the $l$-th and the $(l+1)$-th layers:

\begin{equation}
    y^{(l)} = \sum_{o^{(l,j)}\in\mathcal{O}^{(l)}} \frac{\exp(\alpha^{(l,j)}/T_\alpha)}{\sum_{\alpha^{(l,k)}\in\boldsymbol{\alpha}^{(l)}}\exp(\alpha^{(l,k)}/T_\alpha)} \cdot o^{(l,j)}(x^{(l)}),
\end{equation}
where $T_\alpha$ is a temperature parameter that controls the softness of the operator selection, $y^{(l)}$ is the weighted sum of outputs across all candidate operations at the $l$-th layer and $\boldsymbol{\alpha}$ is the set of all candidate operation weights of all layers.

Additionally, we employ residual connections between the adjacent layers, formulated as:
\begin{equation}
    x^{(l+1)}=x^{(l)}+y^{(l)},
\end{equation}
where $x^{(l+1)}$ represents the input to the $(l+1)$-th layer.

\textcolor{black}{In our KDL-DARTS framework, the losses are tailored to find a compact yet powerful semantic encoder. The training loss, $\mathcal{L}_{\text{train}}$, which is our KD objective in Eq.~\eqref{eq9}, guides the optimization of the encoder's weights ($\boldsymbol{\theta}$) to learn the teacher's semantic representations. Concurrently, the validation loss, $\mathcal{L}_{\text{val}}$, based on our performance-complexity objective (Eq.~\eqref{eq4}), evaluates the quality of the architecture ($\boldsymbol{\alpha}$). This bilevel formulation ensures the inner-loop optimization finds the optimal weights for any given architecture:}
\begin{equation}
\label{eq12}
    \boldsymbol{\theta}^*(\boldsymbol{\alpha}^*) = \arg\min_{\boldsymbol{\theta}} \mathcal{L}_{\text{train}}(\boldsymbol{\theta}, \boldsymbol{\alpha}^*).
\end{equation}



This formulation naturally leads to a bilevel optimization problem, in which $\boldsymbol{\alpha}$ serves as the outer-level optimization variable, and $\boldsymbol{\theta}$ as the inner-level optimization variable:

\begin{equation}
\label{eq13}
 \begin{aligned}
\min_{\boldsymbol{\alpha}}\quad &\mathcal{L}_{\text{val}}\left(\boldsymbol{\theta}^*(\boldsymbol{\alpha}), (\boldsymbol{\alpha})\right) \\
\text{s.t.}\quad& \boldsymbol{\theta}^*(\boldsymbol{\alpha}) = \arg\min_{\boldsymbol{\theta}}\mathcal{L}_{\text{train}}(\boldsymbol{\theta}, \boldsymbol{\alpha}).
\end{aligned}   
\end{equation}

\begin{algorithm}[!t]
\small
\caption{KDL-DARTS}
\label{alg:kdl_darts}
\begin{algorithmic}[1]
    \Require Teacher network $f_{\boldsymbol{\theta}_{\text{t}}}$, Training data $\mathcal{D}_{\text{train}}$, Validation data $\mathcal{D}_{\text{val}}$, Candidate operations $\mathcal{O}^{(l)}$ for layers $l=1...L$, Regularization $\lambda_{\mathcal{J}} > 0$, Learning rates $\eta_{\alpha}$, $\eta_{\theta}$, Step size $\xi$, Number of operations $k$
    \State Initialize architecture parameters $\boldsymbol{\alpha}$ and student network weights $\boldsymbol{\theta}$
    \State Pre-compute penalty factors $\boldsymbol{\beta}$

    \While{\textit{not converged}}
        \State Sample a mini-batch from $\mathcal{D}_{\text{val}}$
        \State Compute gradient $\boldsymbol{g}_{\boldsymbol{\alpha}}$ using approximate gradients and update $\boldsymbol{\alpha}$
        \State Sample a mini-batch from $\mathcal{D}_{\text{train}}$
        \State Compute gradient $\boldsymbol{g}_{\boldsymbol{\theta}}$ and update $\boldsymbol{\theta}$
    \EndWhile

    \State Initialize final architecture $a^* = \emptyset$
    \For{$l = 1$ to $L$}
        \State Compute selection metrics $\boldsymbol{\alpha}^{(l)} \circ (1 - \boldsymbol{\beta}^{(l)})$ and select top $k$ operations for layer $l$
        \State Add selected operations to $a^*$
    \EndFor
    \State \Return Final architecture $a^*$
\end{algorithmic}
\end{algorithm}

\textcolor{black}{Different from the original DARTS, the primary goal of KDL-DARTS is to discover a lightweight semantic encoder architecture capable of effectively learning distilled semantic knowledge. To explicitly encourage a structure suitable for resource-constrained edge devices in our SC system, we integrate an additional regularization term into the validation loss.} Specifically, we introduce a penalty factor set $\boldsymbol{\beta}^{(l)}=\{\beta^{(l,j)}\}\subsetneq \boldsymbol{\beta}$, where $\beta^{(l,j)}$ corresponds to the penalty of the $j$-th candidate in the $l$-th layer. $\boldsymbol{\beta}$ collectively denotes the penalty factors across all layers. Formally, each penalty factor is computed as:

\begin{equation}
    \beta^{(l,j)}= \frac{\exp(|o^{(l,j)}|/T_\beta)}{\sum_{o^{(l,k)}\in\mathcal{O}^{(l)}}\exp(|o^{(l,k)}|/T_\beta)},
\end{equation}
where $T_\beta$ is a temperature parameter about penalty factor and $|o^{(l,k)}|$ represents the number of parameters for operation $o^{(l,k)}$.

The regularization terms for encouraging lightweight architectures in the $l$-th layer is formulated as:
\begin{equation}
    \mathcal{J}^{(l)} = \sum_{j}{\beta^{(l,j)}\cdot \alpha^{(l,j)}},
\end{equation}
where $\mathcal{J}^{(l)}$ represents the regularization for lightweight operation selection in the $l$-th layer.

Therefore, the bilevel optimization problem in (\ref{eq13}) is further modified to explicitly incorporate model complexity constrains as:
\begin{equation}
\label{eq14}
 \begin{aligned}
\min_{\boldsymbol{\alpha}}\quad &\mathcal{L}_{\text{val}}\left(\boldsymbol{\theta}^*(\boldsymbol{\alpha}), (\boldsymbol{\alpha})\right) +\lambda_{\mathcal{J}}\sum_l{\mathcal{J}^{(l)}} \\
\text{s.t.}\quad& \boldsymbol{\theta}^*(\boldsymbol{\alpha}) = \arg\min_{\boldsymbol{\theta}}\mathcal{L}_{\text{train}}(\boldsymbol{\theta}, \boldsymbol{\alpha}),
\end{aligned}   
\end{equation}
where $\lambda_{\mathcal{J}}$ is a positive hyperparameter controlling the relative contribution of the complexity regularization term.


Corresponding to the optimization objective described in (\ref{eq4}), we use the negative validation loss $(-\mathcal{L}_{\text{val}})$ as the measure of model performance and the complexity regularization term $(\sum_l{\mathcal{J}^{(l)}})$ to quantify model complexity. Consequently, the original optimization problem formulated in (\ref{eq5}) can be effectively addressed by solving the bilevel optimization problem defined in (\ref{eq14}).

The introduced regularization term plays an essential role in guiding the architecture search towards a lightweight structure. In particular, during the backward optimization step, the gradient of the regularization term with respect to the architecture parameters $\boldsymbol{\alpha}$ can be expressed as follows:

\begin{equation}
\frac{\partial \mathcal{J}^{(l)}}{\partial \alpha^{(l,j)}} = \frac{\partial \sum_j{\beta^{(l,j)}\cdot\alpha^{(l,j)}}}{\partial \alpha^{(l,j)}}=\beta^{(l,j)}.
\end{equation}

Since the penalty factor $\beta^{(l,j)}$ is independent of the candidate operation weights $\alpha^{(l,j)}$, the partial derivative $\frac{\partial \mathcal{J}^{(l)}}{\partial \alpha^{(l,j)}}$ equals the penalty factor $\beta^{(l,j)}$ itself. Consequently, the magnitude of this derivative is directly proportional to the parameter complexity of the candidate operation $o^{(l,j)}$. As a result, candidate operations with larger numbers of parameters yield higher penalty values and thus produce larger positive gradients. During gradient descent, these larger positive gradients suppress the associated $\alpha^{(l,j)}$ values, effectively reducing the normalized operation weights $w^{(l,j)}$ of computationally expensive operations.

Consequently, as the optimization progresses, the architecture parameters $\boldsymbol{\alpha}$ dynamically adjust to systematically favor candidate operations with fewer parameters. This adaptive process naturally steers the architecture search toward selecting more compact and computationally efficient operations. Ultimately, by incorporating the regularization term into the bilevel optimization framework, the KDL-DARTS algorithm inherently guides the search toward discovering lightweight architectures, effectively balancing high performance achieved through knowledge distillation and desirable computational efficiency of the selected architectures.

For performing an efficient search, we also adopt the approximation scheme used in DARTS:
\begin{equation}\label{eq21}
    \begin{aligned}
    &\nabla_{\boldsymbol{\alpha}} \mathcal{L}_{\text{val}}(\boldsymbol{\theta}^*(\boldsymbol{\alpha}), \boldsymbol{\alpha}) \\
    \approx &\nabla_{\boldsymbol{\alpha}} \mathcal{L}_{\text{val}}(\boldsymbol{\theta} - \xi \nabla_{\boldsymbol{\theta}} \mathcal{L}_{\text{train}}(\boldsymbol{\theta}, \boldsymbol{\alpha}), \boldsymbol{\alpha}),
    \end{aligned}
\end{equation}
where $\xi$ is the learning rate for a step of inner optimization. \textcolor{black}{The computational complexity of KDL-DARTS is dominated by solving the bilevel objective in Eq.~\eqref{eq14}. Following the efficient methodology of DARTS~\cite{liu2018darts}, we use a first-order approximation in Eq.~\eqref{eq21} for the validation loss gradient. Since the gradient of our proposed complexity regularization is computationally inexpensive ($\mathcal{O}(|\boldsymbol{\alpha}|)$), the overall complexity of one optimization step remains $\mathcal{O}(|\boldsymbol{\alpha}| + |\boldsymbol{\theta}|)$.}

Upon completion of the training process, we obtain the optimal architecture parameters $\boldsymbol{\alpha}^*$ and the corresponding model parameters $\boldsymbol{\theta}^*$. The final selection of candidate operations is determined jointly by $\boldsymbol{\alpha}^*$ and the penalty factors $\boldsymbol{\beta}^*$. Specifically, we select the top-$k$ operations based on the combined metric $\alpha^{(l,j)}\cdot (1 - \beta^{(l,j)})$, as illustrated in Fig. \ref{fig:KDL-DARTS}, thereby deriving the optimal architecture $a^*$. Here, the term $(1 - \beta^{(l,j)})$ is adopted instead of $\beta^{(l,j)}$ to effectively penalize candidate operations with larger parameter quantities. Unlike the original DARTS approach, where the final model parameters $\boldsymbol{\theta}^*$ are directly employed following the optimization process, our proposed KDL-DARTS framework solely yields the optimal architecture $a_i^*$ of the $i$-th semantic encoder. In the following sections, we explore how the semantic representation capability of a large-scale model can be effectively transferred to the semantic encoder while maintaining robustness.

\begin{figure*}[t!]
    \centering
    \includegraphics[width=0.9\linewidth]{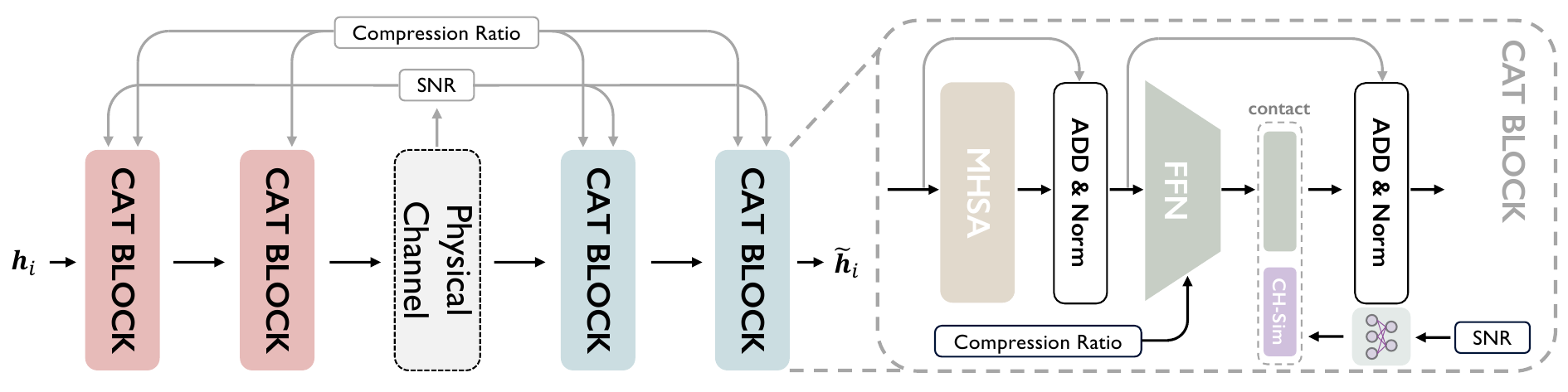}
    \caption{The architecture of proposed Channel Aware Transformer}
    \label{fig:CAT}
\end{figure*}

\subsection{Channel-Aware Transformer}

As formulated in (\ref{eq9}), our training objective comprises two components: a KD loss and a task-specific loss that incorporates channel noise. Here, the KD loss is designed to transfer semantic knowledge from the teacher model to the student model, while the task-specific loss aims to enhance robustness against channel noise. 

In the RKD-SC framework, we propose a RKD algorithm. Unlike conventional KD approaches, RKD not only minimizes the difference between the original student representation $\boldsymbol{h}_i$ and the teacher representation $\boldsymbol{h}_i^\text{Tea}$, but it also minimizes the discrepancy between the noisy student representation $\boldsymbol{\tilde{h}}_i$ and the teacher representation $\boldsymbol{h}_i^\text{Tea}$ to further minimize $\mathcal{L}_{\text{task}}(\boldsymbol{\hat{y},\boldsymbol{y}}\mid \boldsymbol{n})$ in (\ref{eq9}). Directly minimizing $\mathcal{L}_\text{KD}(\boldsymbol{\tilde{h}}_i, \boldsymbol{h}_i^\text{Tea})$ without constraints would allow the transition from $\boldsymbol{\tilde{h}}_i$ to $\boldsymbol{h}_i$ to remain purely random and uncontrolled. To address this, we introduce a CAT module designed to fuse channel-specific semantic features, thereby aiding $\boldsymbol{\tilde{h}}_i$ in effectively approximating $\boldsymbol{h}_i^\text{Tea}$. 

The architecture of CAT is shown in Fig. \ref{fig:CAT}. Both the encoder and decoder of CAT consist of CAT blocks, which are variants of the transformer encoder block \cite{vaswani2023attentionneed}. Unlike the conventional transformer encoder block, the feed-forward network (FFN) within the CAT block has an output dimension smaller than its input dimension. This design choice yields a compact semantic representation, thus reducing the bandwidth requirement.

Due to the constraint that the input and output dimensions of the multi-head self-attention (MHSA) module must match those of the FFN, the channel-specific semantic information derived from the signal-to-noise ratio (SNR) is concatenated with the output of the FFN within the CAT block to compensate for the dimension reduction.  In CAT, a small dense network is employed to transform the SNR into a more fine-grained channel-specific semantic representation.  Notably, the final CAT block in the encoder directly transmits its output to the channel without concatenation, further optimizing bandwidth utilization.

The output dimension of the FFN is governed by a hyperparameter termed the compression ratio which can be calculated as follow:
\begin{equation}
    \text{compression ratio}=1-\frac{\text{dimension of compact features}}{\text{dimension of origin features}}.
\end{equation}

Setting a higher compression ratio results in fewer transmission symbols, significantly conserving resources which also leads to a higher degree of fusion with channel-specific semantic features, potentially causing some loss of the original source information. Conversely, setting a lower compression ratio increases the number of transmission symbols, thus consuming more resources but facilitating greater preservation and incorporation of source semantic features. This can enhance the CAT output's ability to align closely with the semantic representation of a larger-scale model.


The CAT is specifically designed to operate under diverse channel conditions by learning to adaptively fuse channel-specific semantic features. 
Furthermore, guided by the teacher model through distillation, the CAT facilitates the transformation of the semantic features $\boldsymbol{h}_i$, which in turn assists the final representation $\boldsymbol{\tilde{h}}_i$ in effectively approximating the teacher's target features $\boldsymbol{h}_i^\text{Tea}$. 
The methodology for training the CAT module will be elaborated upon below.

 \begin{figure*}[t]
    \centering
    \includegraphics[width=0.85\linewidth]{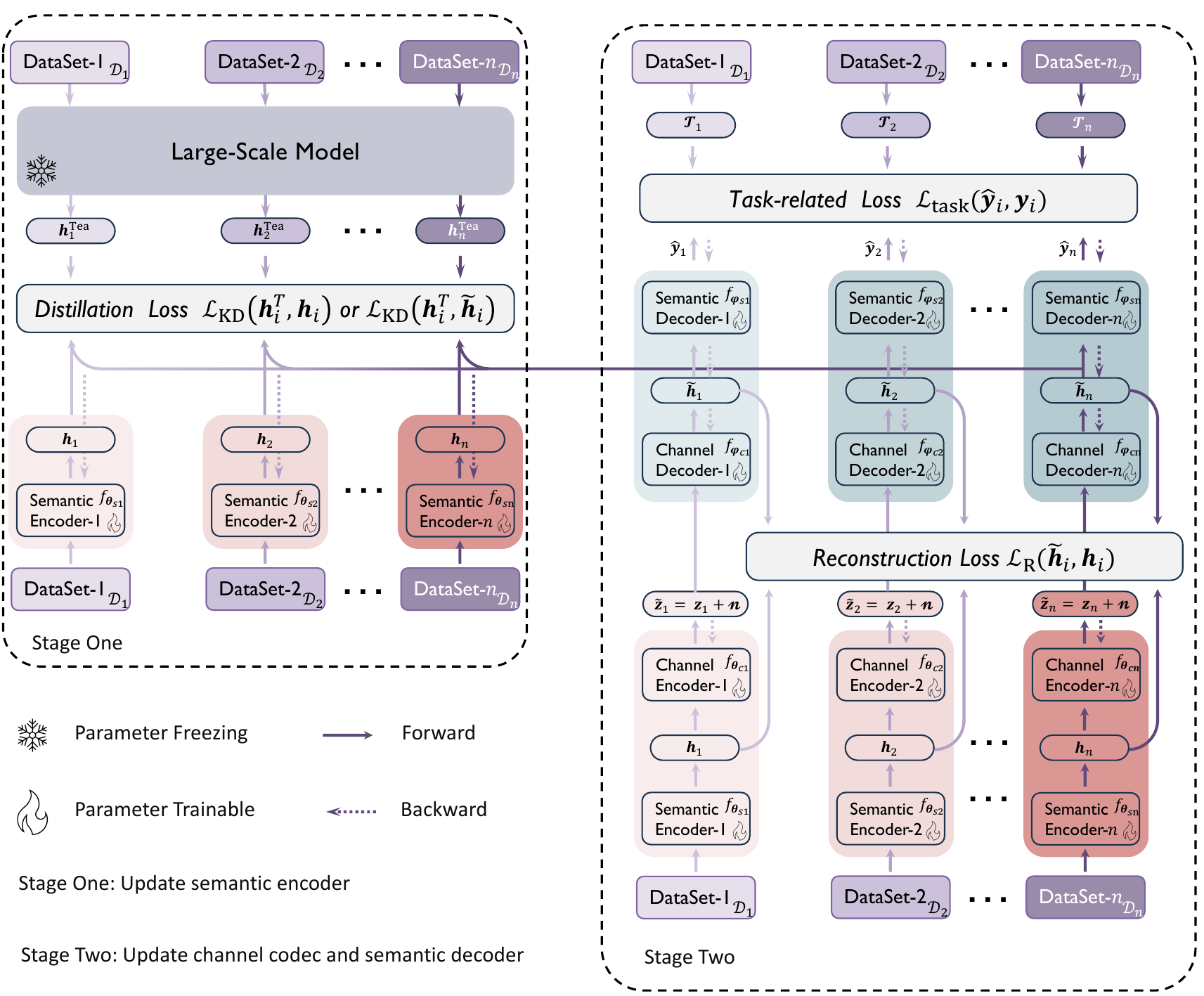}
    \caption{Overall stages of the proposed RKD algorithm: Stage One: Each compact semantic encoder is independently trained, focusing on its designated mission scenario using the corresponding dataset. Stage Two: The channel codec constructed by the CAT is primarily trained with semantic codec jointly.}
    \label{fig:train-stages}
    \vspace{-8pt}
\end{figure*}
\vspace{-8pt}
\subsection{Robust Knowledge Distillation}

\begin{algorithm}[t]
\small
\caption{Two-Stage RKD Algorithmic}
\label{alg:RKD_refined}
\begin{algorithmic}[1]
\Require Teacher model $f_{\boldsymbol{\theta}_\text{t}}$, mission datasets $\mathcal{D}_i$, epochs $E_1, E_2$, batch size $B$, learning rates $\eta_1, \eta_2$, loss weights $\lambda_{\text{KD}}, \lambda_{\text{RE}}, \lambda_{\text{task}}$, optimizer, fixed decoder $f_{\boldsymbol{\varphi}_s}$
\State Initialize parameters for all encoders and decoders.
\Statex

\Statex \textbf{Stage 1: Semantic Encoder Distillation}
\For{$i = 1$ to $n$}
    \For{epoch = 1 to $E_1$}
        \For{each batch $\{(\vect{I}_i^{(k)}, \mathcal{T}_i^{(k)})\}$}
            \State Compute teacher features $\vect{h}_{i,k}^T = f_{\text{teacher}}(\vect{I}_i^{(k)})$
            \State Compute student features $\vect{h}_{i,k} = f_{\boldsymbol{\theta}_{si}}(\vect{I}_i^{(k)})$
            \State Compute distillation loss $\mathcal{L}_{\text{KD}}$
            \State Update encoder parameters: $\boldsymbol{\theta}_{si} \leftarrow \text{Optimize}$
        \EndFor
    \EndFor
    \State Store $\boldsymbol{\theta}_{si}^*$
\EndFor
\Statex

\Statex \textbf{Stage 2: Joint Training Semantic Codec and Channel Codec Optimization}


\For{epoch = 1 to $E_2$}
    \For{each batch $\{(\vect{I}_i^{(k)}, \mathcal{T}_i^{(k)})\}$}
        \State Compute features and losses for small decoder training
        \State Update all parameters: $\boldsymbol{\theta}_{si}, \boldsymbol{\theta}_{ci}, \boldsymbol{\varphi}_c, \boldsymbol{\varphi}_{si}$
    \EndFor
\EndFor

\Statex

\State \textbf{Output:} Optimized parameters 
$\boldsymbol{\theta}_{si}^*, \boldsymbol{\theta}_{ci}^*, \boldsymbol{\varphi}_c^*, \boldsymbol{\varphi}_{si}^*$
\end{algorithmic}
\end{algorithm}

As illustrated in Fig. \ref{fig:train-stages}, the proposed RKD algorithm is a two-stage KD approach designed to address the optimization problem in (\ref{eq9}). In the first stage, a compact semantic encoder is distilled from a large-scale model. Subsequently, in the second stage, the channel codec and the distilled semantic encoder are jointly trained to further enhance the robustness of the SC system.

In stage one, each compact semantic encoder focuses on a specific mission scenario represented by its corresponding dataset. Specifically, for the $i$-th semantic encoder $f_{\boldsymbol{\theta}_{si}}$, there exists a training dataset $\mathcal{D}_i=\{(\boldsymbol{I}_i^{(k)},\mathcal{T}_i^{(k)}), k=1 \cdots M\}$ containing $M$ samples, where $\boldsymbol{I}_i^{(k)}$ denotes the $k$-th sample and $\mathcal{T}_i^{(k)}$ is the corresponding task label. For each training sample $\boldsymbol{I}_i^{(k)}$, the semantic features computed by the large-scale model (teacher model) and the compact semantic encoder (student model) are denoted by $\boldsymbol{h}_{i,k}^\text{Tea}$ and $\boldsymbol{h}_{i,k}$, respectively. The distillation loss is defined as the mean squared error (MSE) between the teacher and student model outputs, formulated as:
\begin{equation}
\label{eq19}
    \ell_{\text{KD}}\left(\boldsymbol{h}_{i,k}^\text{Tea},\boldsymbol{h}_{i,k}\right)=\frac{1}{N}\left\|\boldsymbol{h}_{i,k}^\text{Tea} - \boldsymbol{h}_{i,k} \right\|_2^2,
\end{equation}
where $\ell_{\text{KD}}$ represents the distillation loss function of signal sample, $N$ represents the dimensionality of the semantic feature vectors $\boldsymbol{h}_{i,k}^\text{Tea}$ and $\boldsymbol{h}_{i,k}$ and $\| \cdot \|_2^2$ represents the squared L2 norm of a vector.

The overall distillation loss over the entire dataset is then calculated as:
\vspace{-3pt}
\begin{equation}
\label{eq20}
    \mathcal{L}_{\text{KD}}(\boldsymbol{h}_i^\text{Tea},\boldsymbol{h}_{i})=\frac{1}{M}\sum_{k=1}^{M}{\ell_{\text{KD}}\left(\boldsymbol{h}_{i,k}^\text{Tea},\boldsymbol{h}_{i,k}\right)}.
\end{equation}
\vspace{-3pt}
The optimization objectives of stage one is:
\begin{equation}
    \boldsymbol{\theta}_{si}^*=
    \arg\min_{ \boldsymbol{\theta}_{si}}{\mathcal{L}_{\text{KD}}\left(\boldsymbol{h}_i^\text{Tea}, \boldsymbol{h}_i\right)}.
\end{equation}
\vspace{-3pt}
In this stage, the semantic encoder is optimized by minimizing the distillation loss $\mathcal{L}_{\text{KD}}$ to effectively inherit the knowledge encapsulated within the large-scale model.

In stage two, the channel codec developed by the CAT is primarily trained with semantic codecs jointly. Specifically, for each training sample $I_i^{(k)}$, the sematic encoders $f_{\boldsymbol{\theta}_{s1}}, \cdots, f_{\boldsymbol{\theta}_{sn}}$, the semantic decoders $f_{\boldsymbol{\varphi}_{s1}}, \cdots, f_{\boldsymbol{\varphi}_{sn}}$, the channel encoders $f_{\boldsymbol{\theta}_{c1}}, \cdots, f_{\boldsymbol{\theta}_{cn}}$ and the channel decoder $f_{\boldsymbol{\varphi}_{c1}}, \cdots, f_{\boldsymbol{\varphi}_{cn}}$ are optimized through the following joint loss:
\begin{equation}
\begin{aligned}
    &\mathcal{L}_{\text{joint}}\left(\boldsymbol{h}_i^\text{Tea}, \boldsymbol{h}_i, \tilde{\boldsymbol{h}}_i, \hat{\boldsymbol{y}}_i, \boldsymbol{y}_i\right) = \\
    &\frac{1}{n}\sum_{i=1}^{n}
    \left(
    \lambda_{\text{KD}}\mathcal{L}_{\text{KD}}(\boldsymbol{h}_i^\text{Tea}, \tilde{\boldsymbol{h}}_i ) +
    \lambda_{\text{RE}}\mathcal{L}_{\text{RE}}(\boldsymbol{h}_i, \tilde{\boldsymbol{h}}_i) \right. \\ 
    &\qquad \qquad \quad \qquad \qquad \qquad  \left. + \lambda_{\text{task}}\mathcal{L}_{\text{task}}(\hat{\boldsymbol{y}}_i, \boldsymbol{y}_i)
    \right),
\end{aligned}
\end{equation}
where $\mathcal{L}_\text{RE}$ represents the reconstruction loss that encourages the channel codec to accurately recover the original semantic features despite channel noise. Similar to $\mathcal{L}_{\text{KD}}$, $\mathcal{L}_{\text{RE}}$ is defined as an MSE loss, while $\mathcal{L}_\text{task}$ is task-specific. The hyperparameters $\lambda_\text{KD}$, $\lambda_\text{RE}$ and $\lambda_\text{task}$ control the relative importance of each loss component.

Given that each semantic encoder has already been distilled in stage one, maintaining adequate semantic representational capacity for specific scenarios, they are trained during stage two to enhance the robustness against the channel noise. The optimization objective of stage two is then expressed as:


\begin{equation}
    \label{eq24}
    \begin{aligned}
        \boldsymbol{\theta}_{si}^*, \boldsymbol{\theta}_{ci}^*,\boldsymbol{\varphi}_{ci}^*,\boldsymbol{\varphi}_{si}^*=\\
        \arg\min_{\boldsymbol{\theta}_{si}, \boldsymbol{\theta}_{ci},\boldsymbol{\varphi}_{ci},\boldsymbol{\varphi}_{si}}&{\mathcal{L}_{\text{joint}}\left(\boldsymbol{h}_i^\text{Tea}, \boldsymbol{h}_i, \tilde{\boldsymbol{h}}_i, \hat{\boldsymbol{y}}_i, \boldsymbol{y}_i\right)}\\
        \text{s.t.}\quad& \mathcal{D}= \mathcal{D}_i,
    \end{aligned}
\end{equation}
where $\boldsymbol{\theta}_{si}^*, \boldsymbol{\theta}_{ci}^*,\boldsymbol{\varphi}_{ci}^*,\boldsymbol{\varphi}_{si}^*$ denote the optimal parameters of the $i$-th semantic encoder, the $i$-th semantic encoder, channel encoder, channel decoder, and semantic decoder, respectively. The complete algorithmic process of the RKD algorithm is summarized in Algorithm \ref{alg:RKD_refined}.

Finally, we address the optimization problem presented in (\ref{eq9}). The solution is obtained using our proposed two-stage RKD algorithm, which first distills a large-scale model into compact semantic encoders and subsequently distills the channel codec to enhance noise resistance.
\textcolor{black}{It is important to note that the solution to Eq.~\eqref{eq9} is sub-optimal. Due to the objective function's non-convexity, gradient-based methods converge to a locally optimal solution. The primary complexity is the two-stage training, which, for distilling a ViT-B/16 teacher model in our experiments, required approximately one GPU-day on an NVIDIA RTX 4090. Despite its sub-optimality, this approach proves effective in practice, successfully creating a lightweight yet robust SC system that balances performance with computational cost.}

\section{Simulation Results and Analysis}\label{S4}
We conduct extensive simulations to validate the performance of the proposed RKD-SC framework and analyze its various properties.

\subsection{Simulation Setup}
We consider three transmitters that must deal with image classification tasks of different difficulty, performed on three benchmark datasets: CIFAR10, CIFAR100 \cite{krizhevsky2009learning}, and Tiny-ImageNet (a subset of the ImageNet dataset \cite{5206848}, referred to as ImageNet in this article), respectively. The cross-entropy (CE) loss function is used as the training objective for these classification tasks.

\subsubsection{Architecture and Hyperparameter}


\begin{figure*}[tbp]
  \centering
  \begin{subfigure}[b]{0.3\textwidth}
    \includegraphics[width=\textwidth]{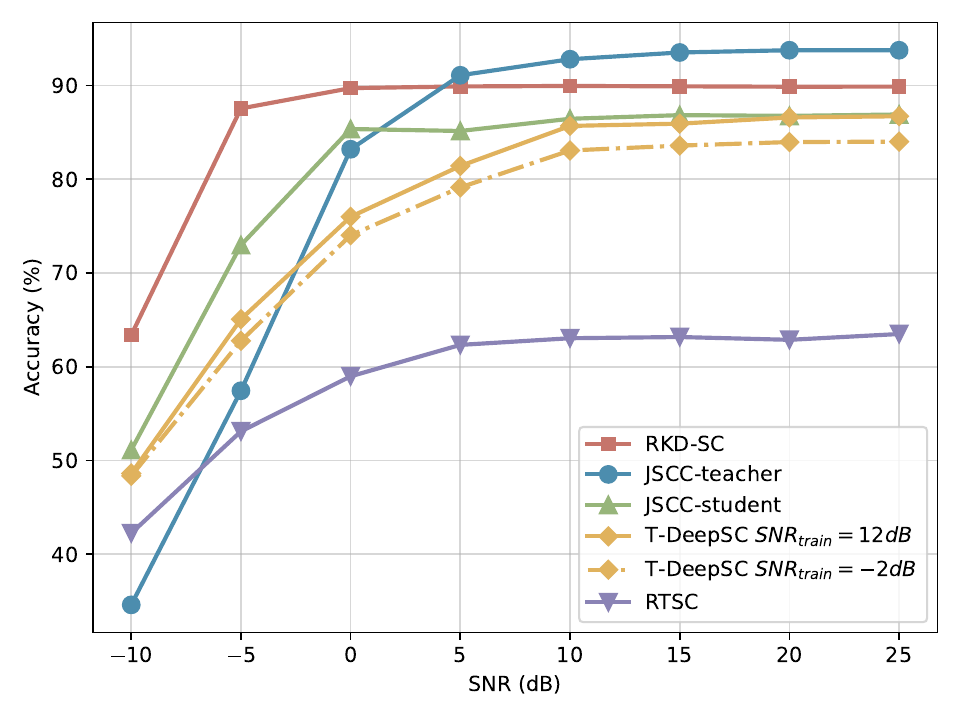}
    \caption{CIFAR10 (AWGN)}
    \label{fig:cifar10_awgn}
  \end{subfigure}
  \hfill
  \begin{subfigure}[b]{0.3\textwidth}
    \includegraphics[width=\textwidth]{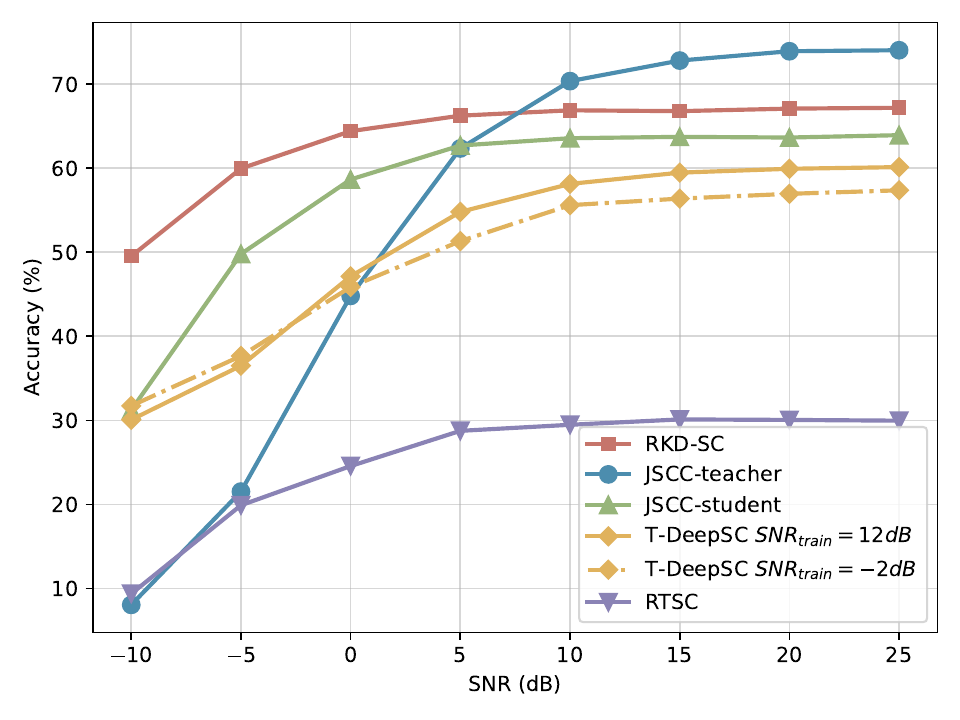}
    \caption{CIFAR100 (AWGN)}
    \label{fig:cifar100_awgn}
  \end{subfigure}
  \hfill
  \begin{subfigure}[b]{0.3\textwidth}
    \includegraphics[width=\textwidth]{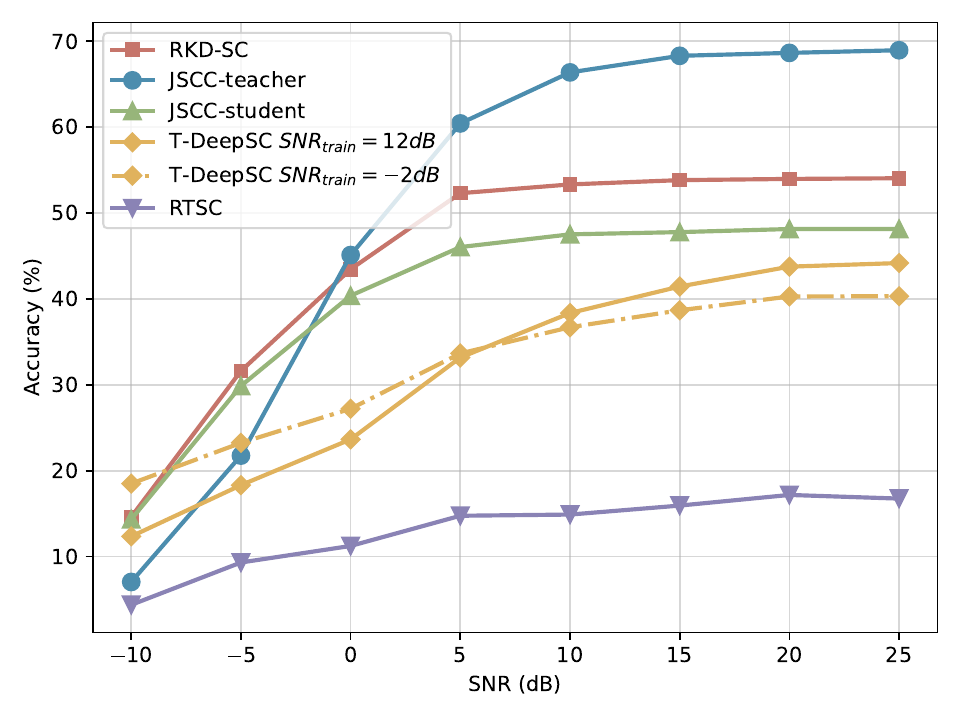}
    \caption{ImageNet (AWGN)}
    \label{fig:imagenet_awgn}
  \end{subfigure}

  \vskip\baselineskip

  \begin{subfigure}[b]{0.3\textwidth}
    \includegraphics[width=\textwidth]{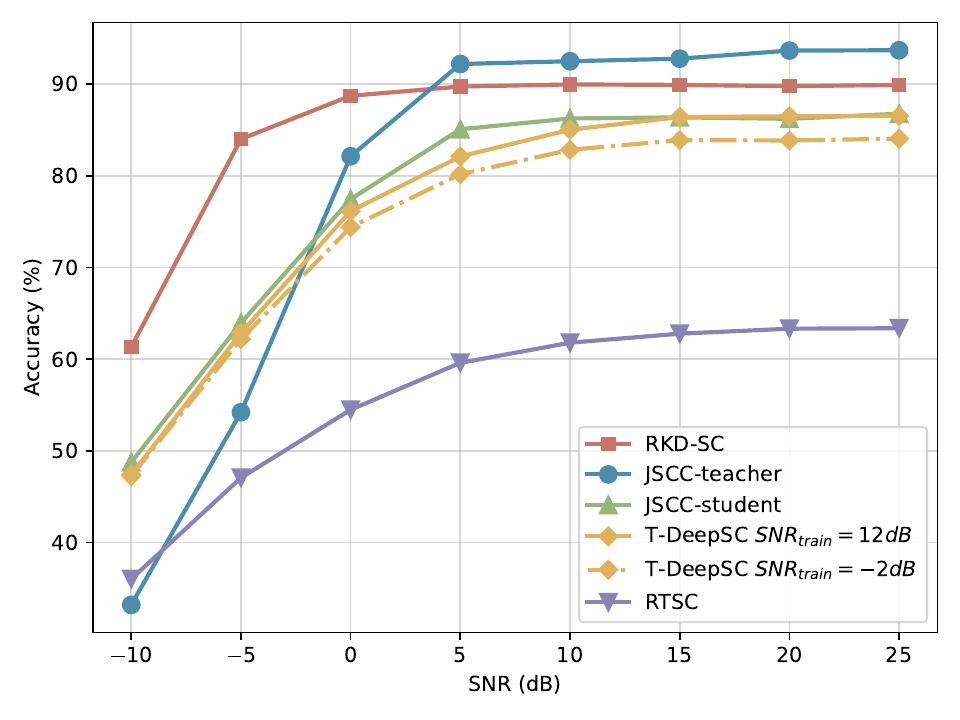}
    \caption{CIFAR10 (Rayleigh Fading)}
    \label{fig:cifar10_rayleigh}
  \end{subfigure}
  \hfill
  \begin{subfigure}[b]{0.3\textwidth}
    \includegraphics[width=\textwidth]{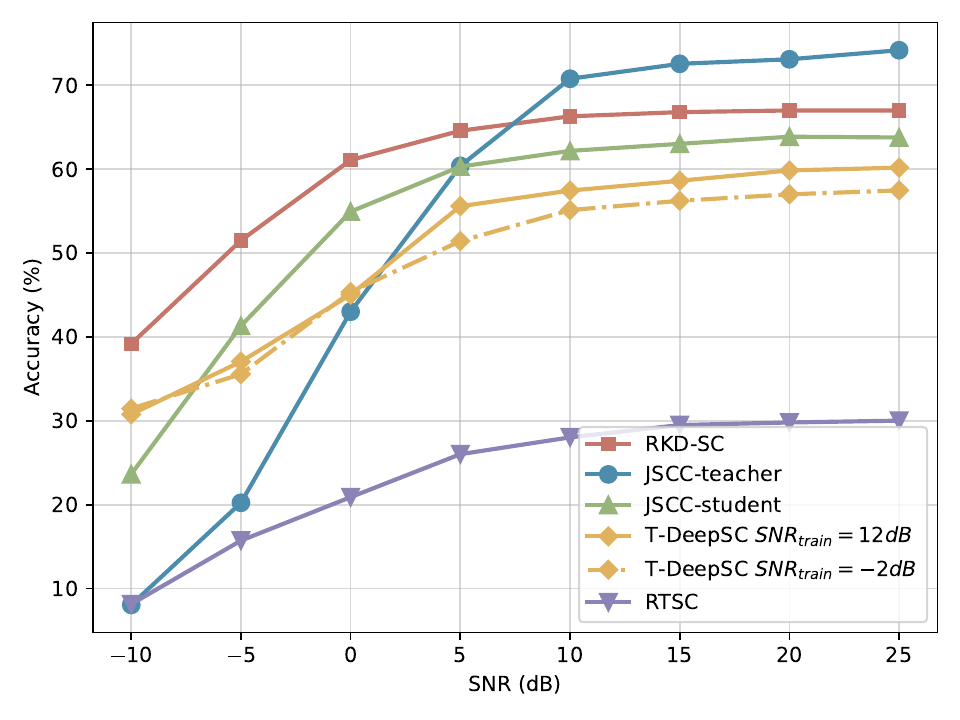}
    \caption{CIFAR100 (Rayleigh Fading)}
    \label{fig:cifar100_rayleigh}
  \end{subfigure}
  \hfill
  \begin{subfigure}[b]{0.3\textwidth}
    \includegraphics[width=\textwidth]{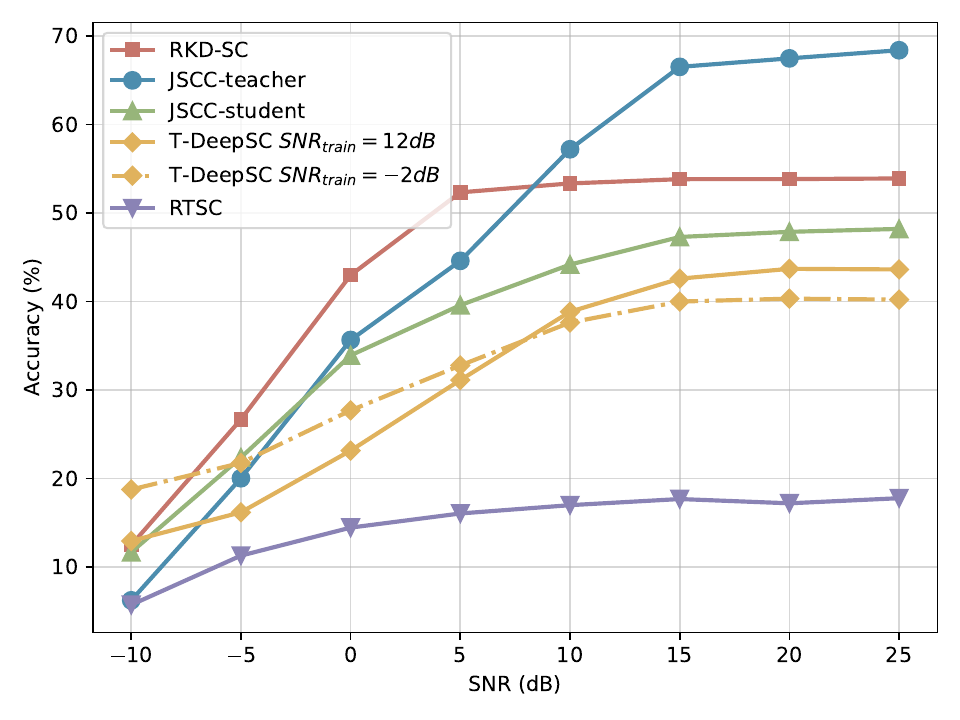}
    \caption{ImageNet (Rayleigh Fading)}
    \label{fig:imagenet_rayleigh}
  \end{subfigure}

  \caption{Comparison of classification accuracy for different datasets under 
  AWGN Channel and Rayleigh Fading Channel.}
  \label{fig:channel_comparison}
\end{figure*}

\begin{table*}[t] 
\centering
\caption{The Architecture of Search Space Overview. Each MixedLayer searches over the number of Bottleneck blocks ($k$).}
\label{tab:nas_search_space_ieee}
\begin{tabular}{@{}lccccc@{}} 
\toprule
\textbf{Stage} & \textbf{Input Size} & \textbf{Output Channels} & \textbf{Stride} & \textbf{Block Type} & \textbf{Configuration / Search Space} \\
\midrule
Input & $224 \times 224 \times 3$ & - & - & - & - \\
\midrule 
Stem Conv1 & $224 \times 224 \times 3$ & 32 & 2 & 3x3 Conv, BN, ReLU & Fixed \\
Stem Conv2 & $112 \times 112 \times 32$ & 32 & 1 & 3x3 Conv, BN, ReLU & Fixed \\
Stem Conv3 & $112 \times 112 \times 32$ & 64 & 1 & 3x3 Conv, BN, ReLU & Fixed \\
Stem Pool & $112 \times 112 \times 64$ & 64 & 2 & 2x2 AvgPool & Fixed \\
\midrule 
Layer 1 & $56 \times 56 \times 64$ & $64~ (16 \times 4)$ & 1 & MixedLayer(Bottleneck) & Depth Search: $k \in \{1..5\}$ blocks \\
Layer 2 & $56 \times 56 \times 128$ & $128~ (32 \times 4)$ & 2 & MixedLayer(Bottleneck) & Depth Search: $k \in \{1..5\}$ blocks \\
Layer 3 & $28 \times 28 \times 256$ & $256~ (64 \times 4)$ & 2 & MixedLayer(Bottleneck) & Depth Search: $k \in \{1..5\}$ blocks \\
Layer 4 & $14 \times 14 \times 512$ & $512~ (128 \times 4)$ & 2 & MixedLayer(Bottleneck) & Depth Search: $k \in \{1..5\}$ blocks \\
\midrule 
Head & $7 \times 7 \times 1024$ & $512$ & - & AttentionPool2d & Fixed (8 heads) \\
\midrule 
Output & - & 512 & - & Feature Vector & - \\
\bottomrule
\end{tabular}
\vspace{-0.5em} 
\end{table*}

\begin{itemize}
\item Search Space $\boldsymbol{\mathcal{A}}$ of Architectures: The compact semantic encoder primarily adopts a residual network structure \cite{he2016deep} integrated with attention pooling. Considering the core objective of KDL-DARTS, which is to identify a lightweight architecture, we simplify the search space $\boldsymbol{\mathcal{A}}$ by focusing exclusively on determining the optimal number of residual blocks within each network layer. Further details are summarized in Table \ref{tab:nas_search_space_ieee}.

\item Architecture of CAT: The CAT block is a standard transformer encoder with 8 heads, 512-dim embeddings, and a 2048-dim feed-forward layer. It includes a single linear downsampling layer for semantic aggregation and a lightweight dense layer (linear upsampling + sigmoid) for channel estimation. The channel encoder uses one CAT block, and the channel decoder uses two.

\item The Teacher Model: The selected teacher model is the Vision Transformer (ViT), originally proposed by \cite{radford2021learningtransferablevisualmodels}. Specifically, we adopt the ViT-B/16 architecture, which comprises approximately 87.85 million parameters.

\item Hyperparameters: In our experiments, KDL-DARTS was configured with 200 search epochs, a complexity regularization weight $\lambda_\mathcal{J}$=0.05, temperature of $\boldsymbol{\alpha}$=1.0, temperature of architecture weights=2.0, and SGD optimizer. We used a CosineAnnealingLR scheduler: the $\boldsymbol{\alpha}$-learning rate decays from 0.025 to $1 \times10^{-4}$, the model learning rate decays from $3\times 10^{-4}$ to $1\times10^{-4}$, and weight decays of $1\times10^{-5}$ ($\boldsymbol{\alpha}$) and $1\times10^{-4}$ (model). The first stage ran for 300 epochs with the learning rate annealed from $5\times10^{-4}$ to $5\times10^{-5}$; The second stage ran for 100 epochs with the learning rate annealed from $5\times 10^{-4}$ to $1\times10^{-5}$. Both stages used CosineAnnealingLR, a training SNR range of 5–20 dB, and CAT compression ratios of 0.8, 0.2 and 0.1 for CIFAR10, CIFAR100 and ImageNet, respectively.


\end{itemize}


\subsubsection{Baselines}
The following baselines are considered.

\begin{itemize}
\item DARTS \cite{liu2018darts}: A differentiable architecture search approach employed within the same architectural search space as KDL-DARTS, detailed in Table \ref{tab:nas_search_space_ieee}.

\item T-DeepSC \cite{qinzhijinUDEEPSC}: A ToSC method leveraging deep learning techniques tailored specifically to targeted applications.

\item RTSC \cite{rtsc}: A real-time SC method utilizing the ViT as its core architecture.

\item JSCC-student: A JSCC method whose encoder shares the same architecture as the semantic encoder implemented in RKD-SC.

\item JSCC-teacher: A JSCC method wherein the encoder directly utilizes the aforementioned teacher model (ViT-B/16).

\end{itemize}

Experiments are conducted on the server equipped with two NVIDIA RTX 4090 GPUs and an Intel\textsuperscript{\textregistered} Core\textsuperscript{\texttrademark} i9-14900KF CPU, operating under Ubuntu 24.04 with CUDA 12.4. The chosen DL framework is PyTorch.

\subsection{Experimental Results}

\begin{figure*}[t]
  \centering
  \begin{subfigure}[t]{\linewidth}
  \centering
    \includegraphics[width=0.87\linewidth]{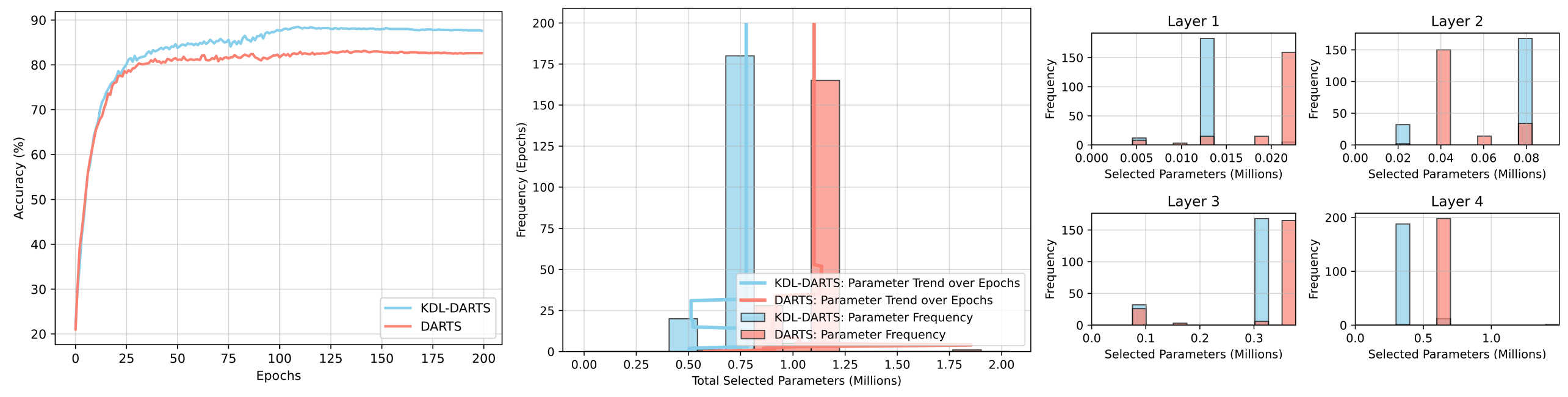}
    \caption{CIFAR10}
    \label{fig:KDL_DARTS_CIFAR10}
  \end{subfigure}

  \vskip\baselineskip

  \begin{subfigure}[t]{\linewidth}
  \centering
    \includegraphics[width=0.87\linewidth]{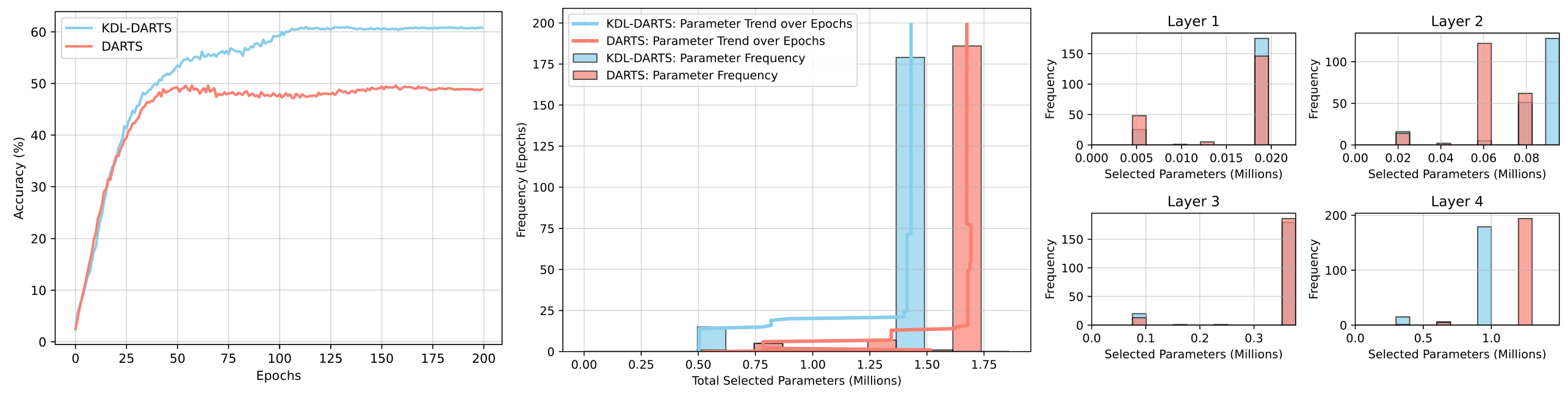}
    \caption{CIFAR100}
    \label{fig:KDL_DARTS_CIFAR100}
  \end{subfigure}

  \vskip\baselineskip

  \begin{subfigure}[t]{\linewidth}
  \centering
    \includegraphics[width=0.87\linewidth]{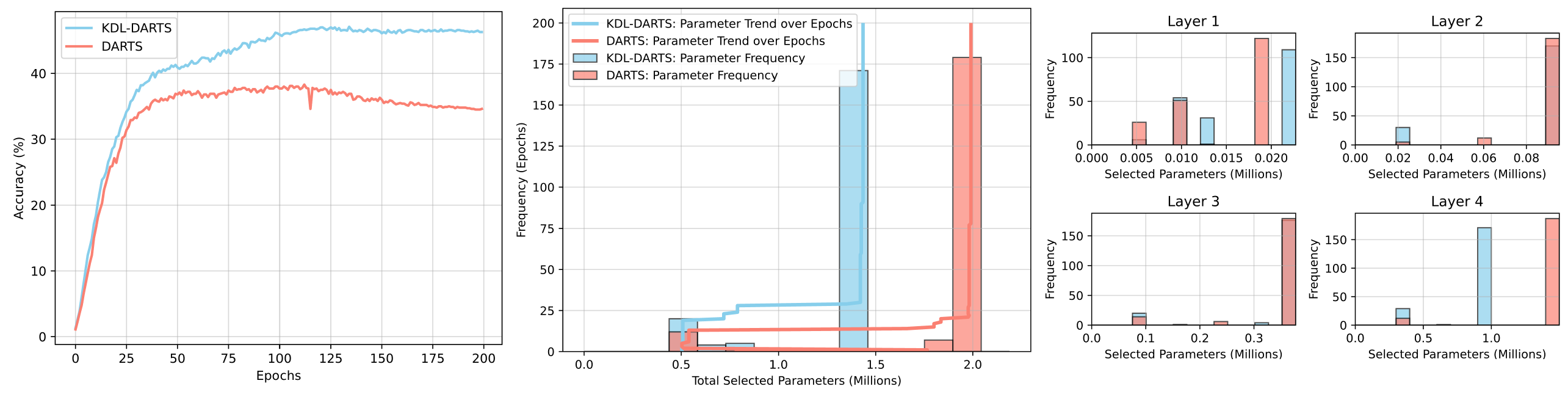}
    \caption{ImageNet}
    \label{fig:KDL_DARTS_ImageNet}
  \end{subfigure}

  \caption{Experimental validation of the proposed KDL-DARTS on (a) CIFAR10, (b) CIFAR100, and (c) ImageNet. In each row: the left plot tracks validation accuracy over epochs; the center plot shows a histogram of total selected parameters (bars) and their evolution over time (line); the right plots detail the per-layer parameter selection frequency across per-layer (Layers 1–4).}
  \label{fig:KDL_DARTS_EXPERIMENTS}
  \vspace{-1em}
\end{figure*}

\subsubsection{Validation of RKD-SC}

Fig. \ref{fig:channel_comparison} presents the results of the RKD-SC framework, which leverages RKD algorithm to fine-tune the architecture identified by KDL-DARTS, compared with several baseline methods. As shown in Fig. \ref{fig:channel_comparison}, under AWGN channels at an SNR of 25 dB, the RKD-SC framework preserves 95.86\%, 90.20\%, and 78.39\% of the performance of the teacher model (labeled as ``JSCC-teacher'' in Fig. \ref{fig:channel_comparison}) on the CIFAR10, CIFAR100, and ImageNet datasets, respectively, while significantly reducing the number of parameters by approximately 94.06\%, 93.27\%, and 93.26\%. Moreover, RKD-SC outperforms the JSCC-student model by over 3.41\%, 5.10\%, and 12.25\% on the three respective datasets. These results indicate that the RKD-SC framework effectively transfers the semantic representation capabilities from the teacher to the compact semantic encoder. Additionally, at an SNR of $-$10 dB, RKD-SC achieves performance gains exceeding 83.12\%, 516.16\%, and 107.51\% compared to the JSCC-teacher on CIFAR10, CIFAR100, and ImageNet datasets, respectively. This demonstrates that the proposed CAT module significantly enhances the robustness against channel noise through the second stage of RKD algorithm within the RKD-SC framework. 

Moreover, the encoder architecture of JSCC-student, which is the same as the RKD-SC framework identified by KDL-DARTS, also exhibits superiority compared to other baseline approaches. As shown in Fig. \ref{fig:channel_comparison}, at an SNR of 0 dB under AWGN channels, JSCC-student achieves higher top-1 accuracy than T-DeepSC trained at an SNR of 12 dB by margins of 9.35\%, 11.53\%, and 16.72\% on CIFAR10, CIFAR100, and ImageNet datasets, respectively, while significantly reducing the parameter count by approximately 81.91\%, 74.04\%, and 73.93\%. These results illustrate that the proposed KDL-DARTS algorithm can help search a lightweight but high performance architecture to complete the specific task.

\subsubsection{Validation of Inference Time} 
Nevertheless, the integration of the CAT module introduces additional processing delay. To quantitatively assess this impact, we conducted further evaluations on the inference time required to encode an image at the transmitter across all investigated methods, as detailed in Table \ref{tab:inference_time_comparison}. In particular, the proposed RKD-SC framework achieves average encoding inference times of 106.21 ms, 127.18 ms, and 130.99 ms on an Internet of Things (IoT\footnotemark \footnotetext{Inference latency on the IoT device (Raspberry Pi 4 B, Broadcom BCM2711) was estimated by scaling the CPU-based inference time according to the ratio of their peak floating-point throughputs. The CPU performance was measured locally at 871.49 GFLOPS, and the Raspberry Pi 4 B peak throughput (32 GFLOPS) was taken from the \href{https://www.cpu-monkey.com/en/cpu-raspberry_pi_4_b_broadcom_bcm2711}{CPU-Monkey database}. Specifically, the Pi inference time $t_{\text{Pi}}$ is approximated as $t_{\text{Pi}} \approx t_{\text{CPU}} \times \frac{\text{GFLOPS}{\text{CPU}}}{\text{GFLOPS}{\text{Pi}}} = t_{\text{CPU}} \times \frac{871.49}{32}$.}) device for the CIFAR10, CIFAR100, and ImageNet datasets, respectively. For comparison, the JSCC-teacher method significantly exceeds these values with an inference time of 1058.86 ms. 
\begin{table*}[t] 
\centering
\caption{Comparison of single image inference times for different methods.} 
\label{tab:inference_time_comparison}
\begin{tabular}{llcccccc}
\toprule
Method & Dataset & Params (M) & GFLOPs & \multicolumn{3}{c}{Inference Time} & Feature Dim \\
\cmidrule(lr){5-7} 
& & & & GPU ($\mu$s) & CPU (ms) & IoT (ms) & \\
\midrule
\multirow{3}{*}{RKD-SC} & CIFAR10 & 5.21 & 0.524 & 30.92 & 3.90 & 106.21 & 102 \\
& CIFAR100 & 5.91 & 0.696 & 40.38 & 4.67 & 127.18 & 409 \\
& ImageNet & 5.92 & 0.711 & 40.41 & 4.81 & 130.99 & 460 \\
\midrule
\multirow{3}{*}{JSCC-student} & CIFAR10 & 1.61 & 0.519 & 21.93 & 3.87 & 105.39 & 512 \\
& CIFAR100 & 2.31 & 0.689 & 28.43 & 4.56 & 124.18 & 512 \\
& ImageNet & 2.32 & 0.704 & 28.57 & 4.61 & 125.54 & 512 \\
\midrule
JSCC-teacher & Dataset Independent & 87.85 & 17.587 & 218.15 & 38.88 & 1058.86 & 512 \\
T-DeepSC & Dataset Independent & 8.90 & 2.266 & 89.34 & 5.53 & 150.60 & 10 (index of KB) \\
RTSC & Dataset Independent & 0.72 & 0.071 & 4.86 & 0.75 & 20.42 & 512 \\
\bottomrule
\end{tabular}
\end{table*}
These empirical results demonstrate that RKD-SC attains superior task performance while incurring only a marginal increase in encoding delay, thus maintaining real-time inference capability. This advantageous balance arises because, although the CAT module possesses a large number of parameters, it processes compact semantic features extracted by the semantic encoder, substantially reducing the required floating point operations (FLOPs) and computational overhead. Furthermore, the CAT module compresses semantic features into more compact representations, reducing the transmitted feature dimensions to 102 for CIFAR10, 409 for CIFAR100, and 460 for ImageNet, instead of the 512 dimensions used by both the JSCC-teacher and JSCC-student models. This reduction shortens transmission delays and offsets the additional processing latency introduced by the CAT module, allowing RKD-SC to maintain overall efficiency.

\begin{figure}[t]
    \centering
    \includegraphics[width=0.8\linewidth]{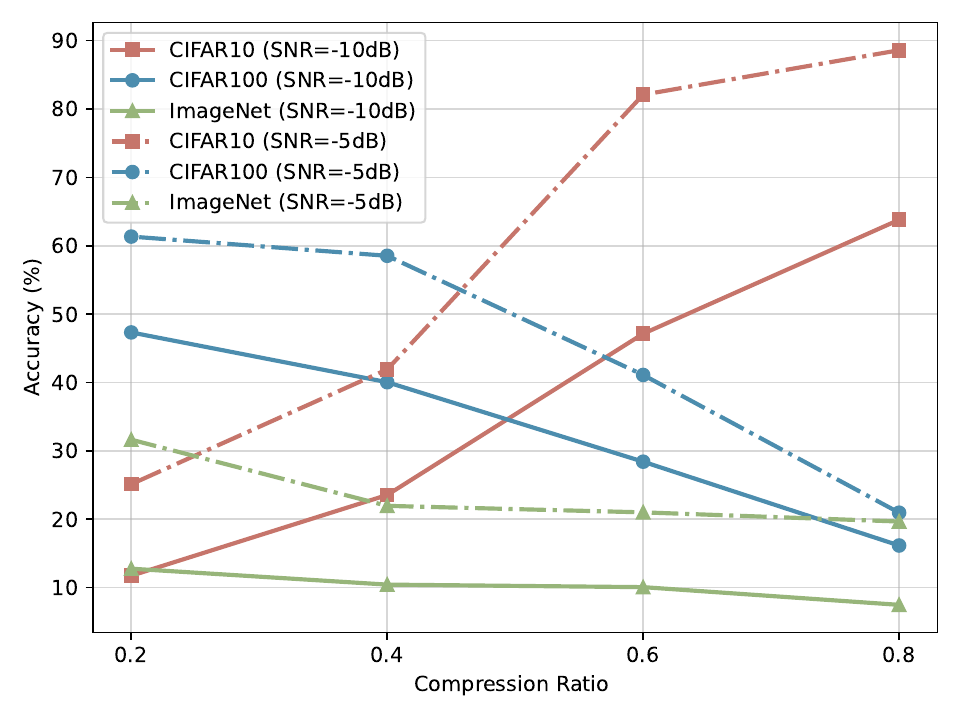}
    \caption{The ablation of compression ratio.}
    \label{fig:ablation}
\end{figure}

\subsubsection{Validation of KDL-DARTS} 

Fig. \ref{fig:KDL_DARTS_EXPERIMENTS} provides results to verify the performance and complexity of the proposed KDL-DARTS algorithm compared with DARTS \cite{liu2018darts}.
From Fig. \ref{fig:KDL_DARTS_EXPERIMENTS}, we observe that KDL-DARTS achieves significant accuracy improvements of over 5\%, 12\%, and 13\% on CIFAR10, CIFAR100, and ImageNet datasets, respectively, as shown in the accuracy-versus-epoch plots (left plots in each row) of Fig. \ref{fig:KDL_DARTS_EXPERIMENTS}. Concurrently, the proposed method reduces the model parameters by approximately 29.4\%, 14.5\%, and 27.9\% for each dataset's selected architecture, as seen in the total parameter analysis plots (center plots in each row) in Fig. \ref{fig:KDL_DARTS_EXPERIMENTS}.

As illustrated in the per-layer parameter histograms (rightmost plots in each row) of Fig. \ref{fig:KDL_DARTS_EXPERIMENTS}, KDL-DARTS tends to select fewer blocks in deeper layers with higher output channel counts, thereby resulting in a lighter architecture compared to DARTS. The lightweight regularization guides the weighting parameters $\boldsymbol{\alpha}$ towards architectures with reduced complexity. Additionally, under the supervision of the teacher model, KDL-DARTS effectively extracts task-relevant semantic features, which reduces the necessity for additional blocks intended to explore deeper semantic information. 

The parameter distribution details for the complete model and individual layers of the final architectures selected by both DARTS and KDL-DARTS are summarized in Table \ref{tab:param_distribution}. These results shown in Fig. \ref{fig:KDL_DARTS_EXPERIMENTS} and Table \ref{tab:param_distribution} collectively demonstrate that the KDL-DARTS approach successfully enhances task-specific performance through the guidance of a high-performing teacher model while also achieving significant parameter efficiency via lightweight regularization and architectural pruning strategies. \textcolor{black}{The efficiency of KDL-DARTS is the same as DARTS the overall cost is with in 1 GPU day on a single NVIDIA RTX 4090.}

\begin{table}[t]
  \centering
  \caption{Comparison of the number of parameters in the final architectures selected by DARTS and KDL-DARTS.}
  \label{tab:param_distribution}
  \begin{tabular}{llrrrrr}
    \toprule
    Dataset      & Method      & \multicolumn{5}{c}{Number of Parameters (M)} \\
    \cmidrule(l){3-7}
                 &             & L1  & L2  & L3  & L4  & Total   \\
    \midrule
    \multirow{2}{*}{CIFAR10}   & DARTS       & 0.227  & 0.420  & 0.377  & 0.660  & 1.101  \\
                               & KDL-DARTS   & 0.014  & 0.078  & 0.307  & 0.379  & 0.777  \\
    \multirow{2}{*}{CIFAR100}  & DARTS       & 0.018  & 0.060  & 0.377  & 0.940  & 1.675  \\
                               & KDL-DARTS   & 0.018  & 0.095  & 0.377  & 1.220  & 1.430  \\
    \multirow{2}{*}{ImageNet}  & DARTS       & 0.023  & 0.095  & 0.377  & 1.500  & 1.990  \\
                               & KDL-DARTS   & 0.018  & 0.095  & 0.377  & 0.940  & 1.435  \\
    \bottomrule
  \end{tabular}
\end{table}

\subsubsection{Ablation Study} Fig. \ref{fig:ablation} illustrates the impact of varying compression ratios within the CAT module. At an SNR of $-10$ dB, for the CIFAR10 dataset, the system's top-1 accuracy significantly improves from 11.73\% to 63.80\% as the compression ratio increases. Conversely, for the CIFAR100 and ImageNet datasets, the top-1 accuracy declines from 47.32\% to 16.15\% and from 12.76\% to 7.47\%, respectively, with an increasing compression ratio. This is because, in CAT, a higher compression ratio corresponds to less preservation of source information and a greater incorporation of channel features. For simpler datasets like CIFAR10, fewer semantic features are sufficient to represent the source information; thus, a higher compression ratio effectively enhances task performance under low SNR conditions by introducing rich channel features. However, for more complex datasets like CIFAR100 and ImageNet, richer semantic representations are essential. Although higher compression ratios introduce additional channel features, the substantial loss of critical source information negatively impacts overall task performance.


\section{Conclusion}\label{S5}

In this paper, we have proposed the RKD-SC framework to effectively leverage the advanced semantic representation capabilities of large-scale models in semantic communication systems while addressing critical challenges related to computational complexity and channel robustness. Within the RKD-SC framework, we have introduced the KDL-DARTS algorithm, which has identified optimal lightweight semantic encoder architectures by incorporating knowledge distillation guidance and complexity regularization into the architecture search process. We have shown that the proposed approach can yield architectures with significantly fewer parameters and improved performance compared to standard DARTS. Moreover, we have shown that the two-stage RKD algorithm, combined with a novel CAT, has effectively transferred knowledge from a large-scale model to the compact student encoder, substantially enhancing the system's robustness against channel noise. Experimental results on CIFAR10, CIFAR100, and ImageNet datasets validated the efficacy of our framework. The results show that RKD-SC can achieve a significant reduction in model parameters while retaining a large fraction of the teacher's performance and demonstrating substantial performance gains, particularly in low SNR regimes, over baseline JSCC methods.



\bibliographystyle{IEEEtran}
\bibliography{Reference}

\begin{thebibliography}{10}
\providecommand{\url}[1]{#1}
\csname url@samestyle\endcsname
\providecommand{\newblock}{\relax}
\providecommand{\bibinfo}[2]{#2}
\providecommand{\BIBentrySTDinterwordspacing}{\spaceskip=0pt\relax}
\providecommand{\BIBentryALTinterwordstretchfactor}{4}
\providecommand{\BIBentryALTinterwordspacing}{\spaceskip=\fontdimen2\font plus
\BIBentryALTinterwordstretchfactor\fontdimen3\font minus \fontdimen4\font\relax}
\providecommand{\BIBforeignlanguage}[2]{{%
\expandafter\ifx\csname l@#1\endcsname\relax
\typeout{** WARNING: IEEEtran.bst: No hyphenation pattern has been}%
\typeout{** loaded for the language `#1'. Using the pattern for}%
\typeout{** the default language instead.}%
\else
\language=\csname l@#1\endcsname
\fi
#2}}
\providecommand{\BIBdecl}{\relax}
\BIBdecl

\bibitem{qinzhijinUDEEPSC}
G.~Zhang, Q.~Hu, Z.~Qin, Y.~Cai, G.~Yu, and X.~Tao, ``A unified multi-task semantic communication system for multimodal data,'' \emph{IEEE Trans. Commun.}, vol.~72, no.~7, pp. 4101--4116, July. 2024.

\bibitem{10929033}
W.~Saad, O.~Hashash, C.~K. Thomas, C.~Chaccour, M.~Debbah, N.~Mandayam, and Z.~Han, ``Artificial general intelligence (agi)-native wireless systems: A journey beyond 6g,'' \emph{Proc. IEEE}, pp. 1--39, March. 2025.

\bibitem{sunlunanJSCC}
L.~Sun, Y.~Yang, M.~Chen, C.~Guo, W.~Saad, and H.~V. Poor, ``Adaptive information bottleneck guided joint source and channel coding for image transmission,'' \emph{IEEE J. Sel. Areas Commun.}, vol.~41, no.~8, pp. 2628--2644, August. 2023.

\bibitem{liuchuanhongOFDM}
C.~Liu, C.~Guo, Y.~Yang, W.~Ni, and T.~Q.~S. Quek, ``Ofdm-based digital semantic communication with importance awareness,'' \emph{IEEE Trans. Commun.}, vol.~72, no.~10, pp. 6301--6315, October. 2024.

\bibitem{10554663}
C.~Chaccour, W.~Saad, M.~Debbah, Z.~Han, and H.~Vincent~Poor, ``Less data, more knowledge: Building next-generation semantic communication networks,'' \emph{IEEE Commun. Surveys Tuts}, vol.~27, no.~1, pp. 37--76, June. 2025.

\bibitem{shannon1949mathematical}
C.~Shannon and W.~Weaver, \emph{The Mathematical Theory of Communication}.\hskip 1em plus 0.5em minus 0.4em\relax University of Illinois Press, 1949.

\bibitem{deepJSCC}
E.~Bourtsoulatze, D.~B. Kurka, and D.~Gündüz, ``Deep joint source-channel coding for wireless image transmission,'' in \emph{ICASSP 2019 - 2019 IEEE International Conference on Acoustics, Speech and Signal Processing (ICASSP)}, May. 2019, pp. 4774--4778.

\bibitem{scalinglaws}
J.~Kaplan, S.~McCandlish, T.~Henighan, T.~B. Brown, B.~Chess, R.~Child, S.~Gray, A.~Radford, J.~Wu, and D.~Amodei, ``Scaling laws for neural language models,'' \emph{arXiv preprint arXiv:2001.08361}, January. 2020.

\bibitem{qin2025lmEmpowerSC}
H.~Xie, Z.~Qin, X.~Tao, and Z.~Han, ``Toward intelligent communications: Large model empowered semantic communications,'' \emph{IEEE Communications Magazine}, vol.~63, no.~1, pp. 69--75, January. 2025.

\bibitem{shahid2025largescaleaitelecomcharting}
\BIBentryALTinterwordspacing
A.~Shahid, A.~Kliks, A.~Al-Tahmeesschi, and et. al, ``Large-scale ai in telecom: Charting the roadmap for innovation, scalability, and enhanced digital experiences,'' March. 2025. [Online]. Available: \url{arXiv preprint arXiv:2503.04184}
\BIBentrySTDinterwordspacing

\bibitem{zhao2023survey}
W.~X. Zhao, K.~Zhou, J.~Li, T.~Tang, X.~Wang, Y.~Hou, Y.~Min, B.~Zhang, J.~Zhang, Z.~Dong \emph{et~al.}, ``A survey of large language models,'' \emph{arXiv preprint arXiv:2303.18223}, March. 2023.

\bibitem{deepseek-r1}
DeepSeek-AI, ``Deepseek-r1: Incentivizing reasoning capability in llms via reinforcement learning,'' \emph{arXiv preprint arXiv:2501.12948}, January. 2025.

\bibitem{Grok3}
\BIBentryALTinterwordspacing
xAI. (2025, February.) Grok 3 beta — the age of reasoning agents. [Online]. Available: \url{https://x.ai/blog/grok-3.}
\BIBentrySTDinterwordspacing

\bibitem{GPT-o3}
\BIBentryALTinterwordspacing
OpenAI. (2025, January.) Openai o3-mini. [Online]. Available: \url{https://openai.com/index/openai-o3-mini/.}
\BIBentrySTDinterwordspacing

\bibitem{farsad2018deeplearningjointsourcechannel}
N.~Farsad, M.~Rao, and A.~Goldsmith, ``Deep learning for joint source-channel coding of text,'' \emph{arXiv preprint arXiv:1802.06832}, February. 2018.

\bibitem{djscc}
E.~Bourtsoulatze, D.~Burth~Kurka, and D.~Gündüz, ``Deep joint source-channel coding for wireless image transmission,'' \emph{IEEE Trans. Cogn. Commun. Netw.}, vol.~5, no.~3, pp. 567--579, September. 2019.

\bibitem{park2025djscc}
J.~Park, Y.~Oh, S.~Kim, and Y.-S. Jeon, ``Joint source-channel coding for channel-adaptive digital semantic communications,'' \emph{IEEE Trans. Cogn. Commun. Netw.}, vol.~11, no.~1, pp. 75--89, February. 2025.

\bibitem{xie2021deep}
H.~Xie, Z.~Qin, G.~Y. Li, and B.-H. Juang, ``Deep learning enabled semantic communication systems,'' \emph{IEEE transactions on signal processing}, vol.~69, pp. 2663--2675, April. 2021.

\bibitem{zhenzi2021DeepSC_S}
Z.~Weng and Z.~Qin, ``Semantic communication systems for speech transmission,'' \emph{IEEE J. Sel. Areas Commun.}, vol.~39, no.~8, pp. 2434--2444, August. 2021.

\bibitem{qinzhijin2022multi_user}
H.~Xie, Z.~Qin, X.~Tao, and K.~B. Letaief, ``Task-oriented multi-user semantic communications,'' \emph{IEEE J. Sel. Areas Commun.}, vol.~40, no.~9, pp. 2584--2597, 2022.

\bibitem{liuchuanhong2024explainable}
C.~Liu, C.~Guo, Y.~Yang, W.~Ni, Y.~Zhou, L.~Li, and T.~Q.~S. Quek, ``Explainable semantic communication for text tasks,'' \emph{IEEE Internet Things J.}, vol.~11, no.~24, pp. 39\,820--39\,833, December. 2024.

\bibitem{fu2024generativeaidriventaskoriented}
Y.~Fu, W.~Cheng, J.~Wang, L.~Yin, and W.~Zhang, ``Generative ai driven task-oriented adaptive semantic communications,'' \emph{arXiv preprint arXiv:2407.11354}, July. 2024.

\bibitem{wang2024largelanguagemodelenabled}
Z.~Wang, L.~Zou, S.~Wei, F.~Liao, J.~Zhuo, H.~Mi, and R.~Lai, ``Large language model enabled semantic communication systems,'' \emph{arXiv preprint arXiv:2407.14112}, July 2024.

\bibitem{ribouh2025largelanguagemodelbasedsemantic}
S.~Ribouh and O.~Saleem, ``Large language model-based semantic communication system for image transmission,'' \emph{arXiv preprint arXiv:2501.12988}, January. 2025.

\bibitem{guoshuaishuai2025}
S.~Guo, Y.~Wang, J.~Ye, A.~Zhang, and K.~Xu, ``Semantic importance-aware communications with semantic correction using large language models,'' \emph{arXiv preprint arXiv:2405.16011}, May. 2024.

\bibitem{yang2025rethinkinggenerativesemanticcommunication}
W.~Yang, Z.~Xiong, S.~Mao, T.~Q.~S. Quek, P.~Zhang, M.~Debbah, and R.~Tafazolli, ``Rethinking generative semantic communication for multi-user systems with large language models,'' \emph{arXiv preprint arXiv:2408.08765}, August. 2024.

\bibitem{xiang2025sceneunderstandingenabledsemantic}
Z.~Xiang, F.~Yu, Q.~Deng, Y.~Li, and Z.~Wan, ``Scene understanding enabled semantic communication with open channel coding,'' \emph{arXiv preprint arXiv:2501.14520}, January. 2025.

\bibitem{hinton2015distillingknowledgeneuralnetwork}
G.~Hinton, O.~Vinyals, and J.~Dean, ``Distilling the knowledge in a neural network,'' \emph{arXiv preprint arXiv:1503.02531}, March. 2015.

\bibitem{RKDSC}
K.~Ding, F.~Liu, Y.~Yang, M.~Chen, and C.~Guo, ``Large scale model enabled semantic communications based on robust knowledge distillation,'' in \emph{GLOBECOM 2024 - 2024 IEEE Global Communications Conference}, December. 2024, pp. 5235--5240.

\bibitem{liu2018darts}
H.~Liu, K.~Simonyan, and Y.~Yang, ``Darts: Differentiable architecture search,'' \emph{arXiv preprint arXiv:1806.09055}, June. 2018.

\bibitem{vaswani2023attentionneed}
A.~Vaswani, N.~Shazeer, N.~Parmar, J.~Uszkoreit, L.~Jones, A.~N. Gomez, L.~Kaiser, and I.~Polosukhin, ``Attention is all you need,'' \emph{arXiv preprint arXiv:1706.03762}, June. 2017.

\bibitem{krizhevsky2009learning}
A.~Krizhevsky, G.~Hinton \emph{et~al.}, ``Learning multiple layers of features from tiny images,'' \emph{University of Toronto Tech. Rep}, vol.~1, January. 2009.

\bibitem{5206848}
J.~Deng, W.~Dong, R.~Socher, L.-J. Li, K.~Li, and L.~Fei-Fei, ``Imagenet: A large-scale hierarchical image database,'' in \emph{2009 IEEE Conference on Computer Vision and Pattern Recognition}, 2009, pp. 248--255.

\bibitem{he2016deep}
K.~He, X.~Zhang, S.~Ren, and J.~Sun, ``Deep residual learning for image recognition,'' in \emph{Proceedings of the IEEE conference on computer vision and pattern recognition}, June. 2016, pp. 770--778.

\bibitem{radford2021learningtransferablevisualmodels}
A.~Radford, J.~W. Kim, C.~Hallacy, A.~Ramesh, G.~Goh, S.~Agarwal, G.~Sastry, A.~Askell, P.~Mishkin, J.~Clark, G.~Krueger, and I.~Sutskever, ``Learning transferable visual models from natural language supervision,'' \emph{arXiv preprint arXiv:2103.00020}, March. 2021.

\bibitem{rtsc}
H.~Yoo, T.~Jung, L.~Dai, S.~Kim, and C.-B. Chae, ``Demo: Real-time semantic communications with a vision transformer,'' in \emph{2022 IEEE International Conference on Communications Workshops (ICC Workshops)}, May. 2022, pp. 1--2.

\end{thebibliography}

\end{document}